# Semi-Implicit Variational Inference

**Mingzhang Yin** [1] **Mingyuan Zhou** [2]

## Abstract

Semi-implicit variational inference (SIVI) is introduced to expand the commonly used analytic variational distribution family, by mixing the variational parameter with a flexible distribution. This mixing distribution can assume any density function, explicit or not, as long as independent random samples can be generated via reparameterization. Not only does SIVI expand the variational family to incorporate highly flexible variational distributions, including implicit ones that have no analytic density functions, but also sandwiches the evidence lower bound (ELBO) between a lower bound and an upper bound, and further derives an asymptotically exact surrogate ELBO that is amenable to optimization via stochastic gradient ascent. With a substantially expanded variational family and a novel optimization algorithm, SIVI is shown to closely match the accuracy of MCMC in inferring the posterior in a variety of Bayesian inference tasks.

## 1. Introduction

Variational inference (VI) is an optimization based method that is widely used for approximate Bayesian inference. It introduces variational distribution $Q$ over the latent variables to approximate the posterior (Jordan et al., 1999), and its stochastic version is scalable to big data (Hoffman et al., 2013). VI updates the parameters of $Q$ to move it closer to the posterior in each iteration, where the closeness is in general measured by the Kullback–Leibler (KL) divergence from the posterior to $Q$, minimizing which is shown to be the same as maximizing the evidence lower bound (ELBO) (Jordan et al., 1999). To make it simple to climb the ELBO to a local optimum, one often takes the mean-field assumption that $Q$ is factorized over the latent variables. The optimization problem is further simplified if each latent variable's distribution is in the same exponential family as its prior, which allows exploiting conditional conjugacy to derive closed-form coordinate-ascent update equations (Bishop & Tipping, 2000; Blei et al., 2017).

Despite its popularity, VI has a well-known issue in underestimating the variance of the posterior, which is often attributed to the mismatch between the representation power of the variational family that $Q$ is restricted to and the complexity of the posterior, and the use of KL divergence, which is asymmetric, to measure how different $Q$ is from the posterior. This issue is often further amplified in mean-field VI (MFVI), due to the factorized assumption on $Q$ that ignores the dependencies between different factorization components (Wainwright et al., 2008; Blei et al., 2017). While there exists a variety of methods that add some structure back to $Q$ to partially restore the dependencies (Saul & Jordan, 1996; Jaakkola & Jordan, 1998; Hoffman & Blei, 2015; Giordano et al., 2015; Tran et al., 2015; 2016; Han et al., 2016; Ranganath et al., 2016; Maaløe et al., 2016; Gregor et al., 2015), it is still necessary for $Q$ to have an analytic probability density function (PDF).

To further expand the variational family that $Q$ belongs to, there has been significant recent interest in defining $Q$ with an implicit model, which makes the PDF of $Q$ become intractable (Huszár, 2017; Mohamed & Lakshminarayanan, 2016; Tran et al., 2017; Li & Turner, 2017; Mescheder et al., 2017; Shi et al., 2017). While using an implicit model could make $Q$ more flexible, it makes it no longer possible to directly computing the log density ratio, as required for evaluating the ELBO. Thus, one often resorts to density ratio estimation, which, however, not only adds an additional level of complexity into each iteration of the optimization, but also is known to be a very difficult problem, especially in high-dimensional settings (Sugiyama et al., 2012).

To well characterize the posterior while maintaining simple optimization, we introduce semi-implicit VI (SIVI) that imposes a mixing distribution on the parameters of the original $Q$ to expand the variational family with a semi-implicit hierarchical construction. The meaning of "semi-implicit" is twofold: 1) the original $Q$ distribution is required to have an analytic PDF, but its mixing distribution is not subject to

[1]Department of Statistics and Data Sciences, [2]Department of IROM, McCombs School of Business, The University of Texas at Austin, Austin TX 78712, USA. Correspondence to: Mingzhang Yin <mzyin@utexas.edu>, Mingyuan Zhou <mingyuan.zhou@mccombs.utexas.edu>.

*Proceedings of the 35<sup>th</sup> International Conference on Machine Learning*, Stockholm, Sweden, PMLR 80, 2018.

such a constraint; and 2) even if both the original $Q$ and its mixing distribution have analytic PDFs, it is common that the marginal of the hierarchy is implicit, that is, having a non-analytic PDF. Our intuition behind SIVI is that even if this marginal is not tractable, its density can be evaluated with Monte Carlo estimation under the semi-implicit hierarchical construction, an expansion that helps model skewness, kurtosis, multimodality, and other characteristics that are exhibited by the posterior but failed to be captured by the original variational family. For MFVI, a benefit of this expansion is restoring the dependencies between its factorization components, as the resulted $Q$ distribution becomes conditionally independent but marginally dependent.

SIVI makes three major contributions: 1) a reparameterizable implicit distribution can be used as a mixing distribution to effectively expand the richness of the variational family; 2) an analytic conditional $Q$ distribution is used to sidestep the hard problem of density ratio estimation, and is not required to be reparameterizable in conditionally conjugate models; and 3) SIVI sandwiches the ELBO between a lower bound and an upper bound, and derives an asymptotically exact surrogate ELBO that is amenable to direct optimization via stochastic gradient ascent. With a flexible variational family and novel optimization, SIVI bridges the accuracy gap of posterior estimation between VI and Markov chain Monte Carlo (MCMC), which can accurately characterize the posterior using MCMC samples, as will be demonstrated in a variety of Bayesian inference tasks. Code is provided at `https://github.com/mingzhang-yin/SIVI`

## 2. Semi-Implicit Variational Inference

In VI, given observations $\boldsymbol{x}$, latent variables $\boldsymbol{z}$, model likelihood $p(\boldsymbol{x} \,|\, \boldsymbol{z})$, and prior $p(\boldsymbol{z})$, we approximate the posterior $p(\boldsymbol{z} \,|\, \boldsymbol{x})$ with variational distribution $q(\boldsymbol{z} \,|\, \boldsymbol{\psi})$ that is often required to be explicit. We optimize the variational parameter $\boldsymbol{\psi}$ to minimize $\text{KL}(q(\boldsymbol{z} \,|\, \boldsymbol{\psi}) || p(\boldsymbol{z} \,|\, \boldsymbol{x}))$, the KL divergence from $p(\boldsymbol{z} \,|\, \boldsymbol{x})$ to $q(\boldsymbol{z} \,|\, \boldsymbol{\psi})$. Since one may show $\log p(\boldsymbol{x}) = \text{ELBO} + \text{KL}(q(\boldsymbol{z} \,|\, \boldsymbol{\psi}) || p(\boldsymbol{z} \,|\, \boldsymbol{x}))$, where

$$\text{ELBO} = -\mathbb{E}_{\boldsymbol{z} \sim q(\boldsymbol{z}|\boldsymbol{\psi})}[\log q(\boldsymbol{z}|\boldsymbol{\psi}) - \log p(\boldsymbol{x}, \boldsymbol{z})], \quad (1)$$

minimizing $\text{KL}(q(\boldsymbol{z} \,|\, \boldsymbol{\psi}) || p(\boldsymbol{z} \,|\, \boldsymbol{x}))$ is equivalent to maximizing the ELBO (Bishop & Tipping, 2000; Blei et al., 2017). Rather than treating $\boldsymbol{\psi}$ as the variational parameter to be inferred, SIVI regards $\boldsymbol{\psi} \sim q(\boldsymbol{\psi})$ as a random variable, as described below. Note that when $q(\boldsymbol{\psi})$ degenerates to a point mass density, SIVI reduces to vanilla VI.

### 2.1. Semi-Implicit Variational Family

Assuming $\boldsymbol{\psi} \sim q_{\boldsymbol{\phi}}(\boldsymbol{\psi})$, where $\boldsymbol{\phi}$ denotes the distribution parameter to be inferred, the semi-implicit variational distribution for $\boldsymbol{z}$ can be defined in a hierarchical manner as

$$\boldsymbol{z} \sim q(\boldsymbol{z} \,|\, \boldsymbol{\psi}), \quad \boldsymbol{\psi} \sim q_{\boldsymbol{\phi}}(\boldsymbol{\psi}).$$

Marginalizing the intermediate variable $\boldsymbol{\psi}$ out, we can view $\boldsymbol{z}$ as a random variable drawn from distribution family $\mathcal{H}$ indexed by variational parameter $\boldsymbol{\phi}$, expressed as

$$\mathcal{H} = \left\{ h_{\boldsymbol{\phi}}(\boldsymbol{z}) : \; h_{\boldsymbol{\phi}}(\boldsymbol{z}) = \int_{\boldsymbol{\psi}} q(\boldsymbol{z} \,|\, \boldsymbol{\psi}) q_{\boldsymbol{\phi}}(\boldsymbol{\psi}) d\boldsymbol{\psi} \right\}.$$

Note $q(\boldsymbol{z} \,|\, \boldsymbol{\psi})$ is required to be explicit, but the mixing distribution $q_{\boldsymbol{\phi}}(\boldsymbol{\psi})$ is allowed to be implicit. Moreover, unless $q_{\boldsymbol{\phi}}(\boldsymbol{\psi})$ is conjugate to $q(\boldsymbol{z} \,|\, \boldsymbol{\psi})$, the marginal $Q$ distribution $h_{\boldsymbol{\phi}}(\boldsymbol{z}) \in \mathcal{H}$ is often implicit. These are the two reasons for referring to the proposed VI as semi-implicit VI (SIVI).

SIVI requires $q(\boldsymbol{z} \,|\, \boldsymbol{\psi})$ to be explicit, and also requires it to either be reparameterizable, which means $\boldsymbol{z} \sim q(\boldsymbol{z} \,|\, \boldsymbol{\psi})$ can be generated by transforming random noise $\boldsymbol{\varepsilon}$ via function $f(\boldsymbol{\varepsilon}, \boldsymbol{\psi})$, or allow the ELBO in (1) to be analytic. Whereas the mixing distribution $q(\boldsymbol{\psi})$ is required to be reparameterizable but not necessarily explicit. In particular, SIVI draws from $q(\boldsymbol{\psi})$ by transforming random noise $\boldsymbol{\epsilon}$ via a deep neural network, which generally leads to an implicit distribution for $q(\boldsymbol{\psi})$ due to a non-invertible transform.

SIVI is related to the hierarchical variational model (Ranganath et al., 2016; Maaløe et al., 2016; Agakov & Barber, 2004) in using a hierarchical variational distribution, but, as discussed below, differs from it in allowing an implicit mixing distribution $q_{\boldsymbol{\phi}}(\boldsymbol{\psi})$ and optimizing the variational parameter via an asymptotically exact surrogate ELBO. Note as long as $q_{\boldsymbol{\phi}}(\boldsymbol{\psi})$ can degenerate to delta function $\delta_{\boldsymbol{\psi}_0}(\boldsymbol{\psi})$ for arbitrary $\boldsymbol{\psi}_0$, the semi-implicit variational family $\mathcal{H}$ is a strict expansion of the original $\mathcal{Q} = \{q(\boldsymbol{z} \,|\, \boldsymbol{\psi}_0)\}$ family, that is, $\mathcal{Q} \subseteq \mathcal{H}$. For MFVI that assumes $q(\boldsymbol{z} \,|\, \boldsymbol{\psi}) = \prod_m q(z_m \,|\, \psi_m)$, this expansion significantly helps restore the dependencies between $z_m$ if $\psi_m$ are not imposed to be independent between each other.

### 2.2. Implicit Mixing Distribution

While restricting $q(\boldsymbol{z} \,|\, \boldsymbol{\psi})$ to be explicit, SIVI introduces a mixing distribution $q_{\boldsymbol{\phi}}(\boldsymbol{\psi})$ to enhance its representation power. In this paper, we construct $q_{\boldsymbol{\phi}}(\boldsymbol{\psi})$ with an implicit distribution that generates its random samples via a stochastic procedure but may not allow a pointwise evaluable PDF. More specifically, an implicit distribution (Mohamed & Lakshminarayanan, 2016; Tran et al., 2017), consisting of a source of randomness $q(\boldsymbol{\epsilon})$ for $\boldsymbol{\epsilon} \in \mathbb{R}^g$ and a deterministic transform $T_{\boldsymbol{\phi}} : \mathbb{R}^g \to \mathbb{R}^d$, can be constructed as $\boldsymbol{\psi} = T_{\boldsymbol{\phi}}(\boldsymbol{\epsilon}), \; \boldsymbol{\epsilon} \sim q(\boldsymbol{\epsilon})$, with PDF

$$q_{\boldsymbol{\phi}}(\boldsymbol{\psi}) = \frac{\partial}{\partial \psi_1} \dots \frac{\partial}{\partial \psi_d} \int_{T_{\boldsymbol{\phi}}(\boldsymbol{\epsilon}) \leq \boldsymbol{\psi}} q(\boldsymbol{\epsilon}) d\boldsymbol{\epsilon}. \quad (2)$$

When $T_{\boldsymbol{\phi}}$ is invertible and the integration is tractable, the PDF of $\boldsymbol{\psi}$ can be calculated with (2), but this is not the case in general and hence $q_{\phi}(\psi)$ is often implicit. When $T_{\phi}(\cdot)$ is chosen as a deep neural network, thanks to its high modeling capacity, $q_{\phi}(\psi)$ can be highly flexible and the dependencies between the elements of $\psi$ can be well captured.

Prevalently used in the study of thermodynamics, ecology, epidemiology, and differential equation systems, implicit distributions have only been recently introduced in VI to parameterize $q(z \mid \psi)$ (Li & Turner, 2017; Mescheder et al., 2017; Huszár, 2017; Tran et al., 2017). Using implicit distributions with intractable PDF increases flexibility but substantially complicates the optimization problem for VI, due to the need to estimate log density ratios involving intractable PDFs, which is particularly challenging in high dimensions (Sugiyama et al., 2012). By contrast, taking a semi-implicit construction, SIVI offers the best of both worlds: constructing a highly flexible variational distribution, without sacrificing the key benefit of VI in converting posterior inference into an optimization problem that is simple to solve. Below we develop a novel optimization algorithm that exploits SIVI's semi-implicit construction.

## 3. Optimization for SIVI

To optimize the variational parameters of SIVI, below we first derive for the ELBO a lower bound, climbing which, however, could drive the mixing distribution $q_{\phi}(\psi)$ towards a point mass density. To prevent degeneracy, we add a nonnegative regularization term, leading to a surrogate ELBO that is asymptotically exact, as can be further tightened by importance reweighting. To sandwich the ELBO, we also derive for the ELBO an upper bound, optimizing which, however, may lead to divergence. We further show that this upper bound can be corrected to a tighter upper bound that monotonically converges from above towards the ELBO.

### 3.1. Lower Bound of ELBO

**Theorem 1** (Cover & Thomas (2012)). *The KL divergence from distribution $p(z)$ to distribution $q(z)$, expressed as $\mathrm{KL}(q(z)||p(z))$, is convex in the pair $(q(z), p(z))$.*

Fixing the distribution $p(z)$ in Theorem 1, KL divergence can be viewed as a convex functional in $q(z)$. As in Appendix A, with Theorem 1 and Jensen's inequality, we have

$$\mathrm{KL}(\mathbb{E}_{\psi} q(z \mid \psi)||p(z)) \leq \mathbb{E}_{\psi}\mathrm{KL}(q(z \mid \psi)||p(z)). \quad (3)$$

Thus, using $h_{\phi}(z) = \mathbb{E}_{\psi \sim q_{\phi}(\psi)} q(z \mid \psi)$ as the variational distribution, SIVI has a lower bound for its ELBO as

$$\begin{aligned} &\underline{\mathcal{L}}(q(z \mid \psi), q_{\phi}(\psi)) = \mathbb{E}_{\psi \sim q_{\phi}(\psi)} \mathbb{E}_{z \sim q(z \mid \psi)} \log \tfrac{p(x,z)}{q(z \mid \psi)} \\ &= -\mathbb{E}_{\psi \sim q_{\phi}(\psi)} \mathrm{KL}(q(z \mid \psi)||p(z|x)) + \log p(x) \\ &\leq -\mathrm{KL}(\mathbb{E}_{\psi \sim q_{\phi}(\psi)} q(z \mid \psi)||p(z|x)) + \log p(x) \\ &= \mathcal{L} = \mathbb{E}_{z \sim h_{\phi}(z)} \log \tfrac{p(x,z)}{h_{\phi}(z)}. \end{aligned} \quad (4)$$

The PDF of $h_{\phi}(z)$ is often intractable, especially if $q_{\phi}(\psi)$ is implicit, prohibiting a Monte Carlo estimation of the ELBO $\mathcal{L}$. By contrast, a Monte Carlo estimation of $\underline{\mathcal{L}}$ only requires $q(z \mid \psi)$ to have an analytic PDF and $q_{\phi}(\psi)$ to be convenient to sample from. It is this separation of evaluation and sampling that allows SIVI to combine an explicit $q(z \mid \psi)$ with an implicit $q_{\phi}(\psi)$ that is as powerful as needed, while maintaining tractable computation.

### 3.2. Degeneracy and Regularization

A direct optimization of the lower bound $\underline{\mathcal{L}}$ in (4), however, can suffer from degeneracy, as shown in the proposition below. All proofs are deferred to Appendix A.

**Proposition 1** (Degeneracy). *Let us denote $\psi^* = \arg\max_{\psi} -\mathbb{E}_{z \sim q(z|\psi)} \log \frac{q(z \mid \psi)}{p(x,z)}$, then*

$$\underline{\mathcal{L}}(q(z \mid \psi), q_{\phi}(\psi)) \leq -\mathbb{E}_{z \sim q(z \mid \psi^*)} \log \frac{q(z|\psi^*)}{p(x,z)},$$

*where the equality is true if and only if $q_{\phi}(\psi) = \delta_{\psi^*}(\psi)$.*

Therefore, if optimizing the variational parameter by climbing $\underline{\mathcal{L}}(q(z \mid \psi), q_{\phi}(\psi))$, without stopping the optimization algorithm early, $q_{\phi}(\psi)$ could converge to a point mass density, making SIVI degenerate to vanilla VI.

To prevent degeneracy, we regularize $\underline{\mathcal{L}}$ by adding

$$B_K = \mathbb{E}_{\psi, \psi^{(1)}, \dots, \psi^{(K)} \sim q_{\phi}(\psi)} \mathrm{KL}(q(z \mid \psi)||\tilde{h}_K(z)), \quad (5)$$
$$\text{where} \quad \tilde{h}_K(z) = \tfrac{q(z \mid \psi) + \sum_{k=1}^{K} q(z \mid \psi^{(k)})}{K+1}.$$

Note that $B_K \geq 0$, with $B_K = 0$ if and only if $K = 0$ or $q_{\phi}(\psi)$ degenerates to a point mass density. Therefore, $\underline{\mathcal{L}}_0 = \underline{\mathcal{L}}$ and maximizing $\underline{\mathcal{L}}_K = \underline{\mathcal{L}} + B_K$ with $K \geq 1$ would encourage positive $B_K$ and drive $q(\psi)$ away from degeneracy. Moreover, as $\lim_{K \to \infty} \tilde{h}_K(z) = \mathbb{E}_{\psi \sim q_{\phi}(\psi)} q(z \mid \psi) = h_{\phi}(z)$ by the strong law of large numbers and

$$\lim_{K \to \infty} B_K = \mathbb{E}_{\psi \sim q_{\phi}(\psi)} \mathrm{KL}(q(z \mid \psi)||h_{\phi}(z)) \quad (6)$$

by interchanging two limiting operations, as discussed in detail in Appendix A, we have the following proposition.

**Proposition 2.** *Suppose $\mathcal{L}$ and $\underline{\mathcal{L}}$ are defined as in (4) and $B_K$ as in (5), the regularized lower bound $\underline{\mathcal{L}}_K = \underline{\mathcal{L}} + B_K$ is an asymptotically exact ELBO that satisfies $\underline{\mathcal{L}}_0 = \underline{\mathcal{L}}$ and $\lim_{K \to \infty} \underline{\mathcal{L}}_K = \mathcal{L}$.*

### 3.3. Upper Bound of ELBO and Correction

Using the concavity of the logarithmic function, we have $\log h_{\phi}(z) \geq \mathbb{E}_{\psi \sim q_{\phi}(\psi)} \log q(z \mid \psi)$, and hence we can obtain an upper bound of SIVI's ELBO as

$$\begin{aligned} \bar{\mathcal{L}}(q(z \mid \psi), q_{\phi}(\psi)) &= \mathbb{E}_{\psi \sim q_{\phi}(\psi)} \mathbb{E}_{z \sim h_{\phi}(z)} \log \tfrac{p(x,z)}{q(z \mid \psi)} \\ &\geq \mathcal{L} = \mathbb{E}_{z \sim h_{\phi}(z)} \log \tfrac{p(x,z)}{h_{\phi}(z)}. \end{aligned} \quad (7)$$

Comparing (4) and (7) shows that the lower bound $\underline{\mathcal{L}}$ and upper bound $\bar{\mathcal{L}}$ only differ from each other in whether the expectation involving $\boldsymbol{z}$ is taken with respect to $q(\boldsymbol{z}\,|\,\boldsymbol{\psi})$ or $h_{\boldsymbol{\phi}}(\boldsymbol{z})$. Moreover, one may show that $\bar{\mathcal{L}} - \underline{\mathcal{L}}$ is equal to

$$\mathbb{E}_{\boldsymbol{\psi}\sim q(\boldsymbol{\psi})}[\mathrm{KL}(q(\boldsymbol{z}\,|\,\boldsymbol{\psi})||h_{\boldsymbol{\phi}}(\boldsymbol{z})) + \mathrm{KL}(h_{\boldsymbol{\phi}}(\boldsymbol{z})||q(\boldsymbol{z}\,|\,\boldsymbol{\psi}))].$$

Since $\bar{\mathcal{L}}$ may not be bounded above by the evidence $\log p(\boldsymbol{x})$ and $\bar{\mathcal{L}} - \underline{\mathcal{L}}$ is not bounded from above, there is no convergence guarantee if maximizing $\bar{\mathcal{L}}$. For this reason, we subtract $\bar{\mathcal{L}}$ by a correction term as

$$A_K = \mathbb{E}_{\boldsymbol{\psi}\sim q_{\boldsymbol{\phi}}(\boldsymbol{\psi})}\mathbb{E}_{\boldsymbol{z}\sim h_{\boldsymbol{\phi}}(\boldsymbol{z})}\mathbb{E}_{\boldsymbol{\psi}^{(1)},\ldots,\boldsymbol{\psi}^{(K)}\sim q_{\boldsymbol{\phi}}(\boldsymbol{\psi})}\Big[ \log\big(\tfrac{1}{K}\textstyle\sum_{k=1}^K q(\boldsymbol{z}\,|\,\boldsymbol{\psi}^{(k)})\big) - \log q(\boldsymbol{z}\,|\,\boldsymbol{\psi})\Big]. \quad (8)$$

As $\mathbb{E}_{\boldsymbol{\psi}^{(1)}\sim q_{\boldsymbol{\phi}}(\boldsymbol{\psi})}\log q(\boldsymbol{z}\,|\,\boldsymbol{\psi}^{(1)}) = \mathbb{E}_{\boldsymbol{\psi}\sim q_{\boldsymbol{\phi}}(\boldsymbol{\psi})}\log q(\boldsymbol{z}\,|\,\boldsymbol{\psi})$, we have $A_1 = 0$. The following proposition shows that the corrected upper bound $\bar{\mathcal{L}}_K = \bar{\mathcal{L}} - A_K$ monotonically converges from above towards the ELBO as $K\to\infty$.

**Proposition 3.** *Suppose $\mathcal{L}$ and $\bar{\mathcal{L}}$ are defined as in (7) and $A_K$ as in (8), then the corrected upper bound $\bar{\mathcal{L}}_K = \bar{\mathcal{L}} - A_K$ monotonically converges from the above towards the ELBO, satisfying $\bar{\mathcal{L}}_1 = \bar{\mathcal{L}}$, $\bar{\mathcal{L}}_{K+1}\le\bar{\mathcal{L}}_K$, and $\lim_{K\to\infty}\bar{\mathcal{L}}_K = \mathcal{L}$.*

The relationship between $\underline{\mathcal{L}}_K = \underline{\mathcal{L}} + B_K$ and $\bar{\mathcal{L}}_K = \bar{\mathcal{L}} - A_K$, two different asymptotically exact ELBOs, can be revealed by rewriting them as

$$\underline{\mathcal{L}}_K = \mathbb{E}_{\boldsymbol{\psi}\sim q_{\boldsymbol{\phi}}(\boldsymbol{\psi})}\mathbb{E}_{\boldsymbol{z}\sim q(\boldsymbol{z}\,|\,\boldsymbol{\psi})}\mathbb{E}_{\boldsymbol{\psi}^{(1)},\ldots,\boldsymbol{\psi}^{(K)}\sim q_{\boldsymbol{\phi}}(\boldsymbol{\psi})} \log\frac{p(\boldsymbol{x},\boldsymbol{z})}{\frac{1}{K+1}\left[q(\boldsymbol{z}\,|\,\boldsymbol{\psi})+\sum_{k=1}^K q(\boldsymbol{z}\,|\,\boldsymbol{\psi}^{(k)})\right]}, \quad (9)$$

$$\bar{\mathcal{L}}_K = \mathbb{E}_{\boldsymbol{\psi}\sim q_{\boldsymbol{\phi}}(\boldsymbol{\psi})}\mathbb{E}_{\boldsymbol{z}\sim q(\boldsymbol{z}\,|\,\boldsymbol{\psi})}\mathbb{E}_{\boldsymbol{\psi}^{(1)},\ldots,\boldsymbol{\psi}^{(K)}\sim q_{\boldsymbol{\phi}}(\boldsymbol{\psi})} \log\frac{p(\boldsymbol{x},\boldsymbol{z})}{\frac{1}{K}\sum_{k=1}^K q(\boldsymbol{z}\,|\,\boldsymbol{\psi}^{(k)})}. \quad (10)$$

Thus $\underline{\mathcal{L}}_K$ differs from $\bar{\mathcal{L}}_K$ in whether $q(\boldsymbol{z}\,|\,\boldsymbol{\psi})$ participates in computing the log density ratio, which is analytic thanks to the semi-implicit construction, inside the expectations. When $K$ is small, using $\underline{\mathcal{L}}_K$ as the surrogate ELBO for optimization is expected to have better numerical stability than using $\bar{\mathcal{L}}_K$, as $\underline{\mathcal{L}}_0 = \underline{\mathcal{L}}$ relates to the ELBO as a lower bound while $\bar{\mathcal{L}}_1 = \bar{\mathcal{L}}$ does as an upper bound, but increasing $K$ quickly diminishes the difference between $\underline{\mathcal{L}}_K$ and $\bar{\mathcal{L}}_K$, which are both asymptotically exact. It is also instructive to note that as $\boldsymbol{z}\sim q(\boldsymbol{z}\,|\,\boldsymbol{\psi})$ is used for Monte Carlo estimation, if assuming $\{q(\boldsymbol{z}\,|\,\boldsymbol{\psi}), q(\boldsymbol{z}\,|\,\boldsymbol{\psi}^{(1)}),\ldots,q(\boldsymbol{z}\,|\,\boldsymbol{\psi}^{(K)})\}$ has a dominant element, then it is most likely that $q(\boldsymbol{z}\,|\,\boldsymbol{\psi})$ dominates all $q(\boldsymbol{z}\,|\,\boldsymbol{\psi}^{(k)})$. Therefore, maximizing $\underline{\mathcal{L}}_K$ in (9) would become almost the same as maximizing $\underline{\mathcal{L}}_0$, which would lead to degeneracy as in Proposition 1, which means $\boldsymbol{\psi} = \boldsymbol{\psi}^{(k)}$ and $q(\boldsymbol{z}\,|\,\boldsymbol{\psi}) = q(\boldsymbol{z}\,|\,\boldsymbol{\psi}^{(k)})$ for all $k$, contradicting the reasoning that $q(\boldsymbol{z}\,|\,\boldsymbol{\psi})$ dominates all $q(\boldsymbol{z}|\boldsymbol{\psi}^{(k)})$.

Using the importance reweighting idea, Burda et al. (2015) provides a lower bound $L^{\tilde{K}} \ge$ ELBO that monotonically converges from below to the evidence $\log p(\boldsymbol{x})$ as $\tilde{K}$ increases. Using the same idea, we may also tighten the asymptotically exact surrogate ELBO in (9) using

$$\underline{\mathcal{L}}_K^{\tilde{K}} = \mathbb{E}_{(\boldsymbol{z}_1,\boldsymbol{\psi}_1),\ldots,(\boldsymbol{z}_{\tilde{K}},\boldsymbol{\psi}_{\tilde{K}})\sim q(\boldsymbol{z}\,|\,\boldsymbol{\psi})q_{\boldsymbol{\phi}}(\boldsymbol{\psi})}\mathbb{E}_{\boldsymbol{\psi}^{(1)},\ldots,\boldsymbol{\psi}^{(K)}\sim q_{\boldsymbol{\phi}}(\boldsymbol{\psi})} \log\frac{1}{\tilde{K}}\sum_{i=1}^{\tilde{K}}\frac{p(\boldsymbol{x},\boldsymbol{z}_i)}{\frac{1}{K+1}\left[q(\boldsymbol{z}_i\,|\,\boldsymbol{\psi}_i)+\sum_{k=1}^K q(\boldsymbol{z}_i\,|\,\boldsymbol{\psi}^{(k)})\right]},$$

for which $\lim_{K\to\infty}\underline{\mathcal{L}}_K^{\tilde{K}} = L^{\tilde{K}} \ge$ ELBO and $\lim_{K,\tilde{K}\to\infty}\underline{\mathcal{L}}_K^{\tilde{K}} = \lim_{\tilde{K}\to\infty}L^{\tilde{K}} = \log p(\boldsymbol{x})$.

Using $\underline{\mathcal{L}}_{K_t}$ as the surrogate ELBO, where $t$ indexes the number of iterations, $K_t\in\{0,1,\ldots\}$, and $K_{t+1}\ge K_t$, we describe the stochastic gradient ascent algorithm to optimize the variational parameter in Algorithm 1, in which we further introduce $\boldsymbol{\xi}$ as the variational parameter of the conditional distribution $q_{\boldsymbol{\xi}}(\boldsymbol{z}\,|\,\boldsymbol{\psi})$ that is not mixed with another distribution. For Monte Carlo estimation in Algorithm 1, we use a single random sample for each $\boldsymbol{\psi}^{(k)}$, $J$ random samples for $\boldsymbol{\psi}$, and a single sample of $\boldsymbol{z}$ for each sample of $\boldsymbol{\psi}$. One may further consult Rainforth et al. (2018) to help select $K$, $J$, and $\tilde{K}$ for SIVI. We denote $\boldsymbol{z} = f(\boldsymbol{\varepsilon},\boldsymbol{\xi},\boldsymbol{\psi})$, $\boldsymbol{\varepsilon}\sim p(\boldsymbol{\varepsilon})$ as the reparameterization for $\boldsymbol{z}\sim q_{\boldsymbol{\xi}}(\boldsymbol{z}\,|\,\boldsymbol{\psi})$. As for $\boldsymbol{\xi}$, if $\boldsymbol{\xi}\neq\emptyset$, one may learn it as in Algorithm 1, set it empirically, or fix it at the value learned by another algorithm such as MFVI. In summary, SIVI constructs a flexible variational distribution by mixing a (potentially) implicit distribution with an explicit one, while maintaining tractable optimization via the use of an asymptotically exact surrogate ELBO.

### 3.4. Score Function Gradient in Conjugate Model

Let's consider the case that $q(\boldsymbol{z}\,|\,\boldsymbol{\psi})$ does not have a simple reparameterization but can benefit from conditional conjugacy. In particular, for a conditionally conjugate exponential family model, MFVI has an analytic ELBO, and its variational distribution can be directly used as the $q_{\boldsymbol{\xi}}(\boldsymbol{z}\,|\,\boldsymbol{\psi})$ of SIVI. As in Appendix A, introducing a density ratio as

$$r_{\boldsymbol{\xi},\boldsymbol{\phi}}(\boldsymbol{z},\boldsymbol{\epsilon},\boldsymbol{\epsilon}^{(1:K)}) = \frac{q_{\boldsymbol{\xi}}(\boldsymbol{z}\,|\,T_{\boldsymbol{\phi}}(\boldsymbol{\epsilon}))}{\frac{q_{\boldsymbol{\xi}}(\boldsymbol{z}\,|\,T_{\boldsymbol{\phi}}(\boldsymbol{\epsilon}))+\sum_{k=1}^K q_{\boldsymbol{\xi}}(\boldsymbol{z}\,|\,T_{\boldsymbol{\phi}}(\boldsymbol{\epsilon}^{(k)}))}{K+1}},$$

we approximate the gradient of $\underline{\mathcal{L}}_K$ with respect to $\boldsymbol{\phi}$ as

$$\nabla_{\boldsymbol{\phi}}\underline{\mathcal{L}}_K \approx \frac{1}{J}\sum_{j=1}^J\Big\{-\nabla_{\boldsymbol{\phi}}\mathbb{E}_{\boldsymbol{z}\sim q_{\boldsymbol{\xi}}(\boldsymbol{z}\,|\,T_{\boldsymbol{\phi}}(\boldsymbol{\epsilon}_j))}[\log\frac{q_{\boldsymbol{\xi}}(\boldsymbol{z}\,|\,T_{\boldsymbol{\phi}}(\boldsymbol{\epsilon}_j))}{p(\boldsymbol{x},\boldsymbol{z})}] + \nabla_{\boldsymbol{\phi}}\log r_{\boldsymbol{\xi},\boldsymbol{\phi}}(\boldsymbol{z}_j,\boldsymbol{\epsilon}_j,\boldsymbol{\epsilon}^{(1:K)}) + [\nabla_{\boldsymbol{\phi}}\log q_{\boldsymbol{\xi}}(\boldsymbol{z}_j\,|\,T_{\boldsymbol{\phi}}(\boldsymbol{\epsilon}_j))]\log r_{\boldsymbol{\xi},\boldsymbol{\phi}}(\boldsymbol{z}_j,\boldsymbol{\epsilon}_j,\boldsymbol{\epsilon}^{(1:K)})\Big\}, \quad (11)$$

where the first summation term is equivalent to the gradient of MFVI's ELBO, both the second and third terms correct the restrcitions of $q_{\boldsymbol{\xi}}(\boldsymbol{z}\,|\,T_{\boldsymbol{\phi}}(\boldsymbol{\epsilon}_j))$, and $\log r_{\boldsymbol{\xi},\boldsymbol{\phi}}(\boldsymbol{z},\boldsymbol{\epsilon},\boldsymbol{\epsilon}^{(1:K)})$ in the third term is expected to be small regardless of convergence, effectively mitigating the variance of score function gradient estimation that is usually high in basic black-box VI; $\nabla_{\boldsymbol{\xi}}\underline{\mathcal{L}}_K$ can be approximated in the same manner. For non-conjugate models, we leave for future study the use of non-reparameterizable $q_{\boldsymbol{\xi}}(\boldsymbol{z}\,|\,\boldsymbol{\psi})$, for which one may apply customized variance reduction techniques (Paisley et al., 2012; Ranganath et al., 2014; Mnih & Gregor, 2014; Ruiz et al., 2016; Kucukelbir et al., 2017; Naesseth et al., 2017).

# 4. Related Work

There exists a wide variety of VI methods that improve on MFVI. Examples include adding dependencies between latent variables (Saul & Jordan, 1996; Hoffman & Blei, 2015), using a mixture of variational distributions (Bishop et al., 1998; Gershman et al., 2012; Salimans & Knowles, 2013; Guo et al., 2016; Miller et al., 2017), introducing a copula to capture the dependencies between univariate marginals (Tran et al., 2015; Han et al., 2016), handling non-conjugacy (Paisley et al., 2012; Titsias & Lázaro-Gredilla, 2014), and constructing a hierarchical variational distribution (Ranganath et al., 2016; Tran et al., 2017).

To increase the expressiveness of the PDF of a random variable, a simple but powerful idea is to transform it with complex deterministic and/or stochastic mappings. One successful application of this idea in VI is constructing the variational distribution with a normalizing flow, which transforms a simple random variable through a sequence of invertible differentiable functions with tractable Jacobians, to deterministically map a simple PDF to a complex one (Rezende & Mohamed, 2015; Kingma et al., 2016; Papamakarios et al., 2017). Normalizing flows help increase the flexibility of VI, but still require the mapping to be deterministic and invertible. Removing both restrictions, there have been several recent attempts to define highly flexible variational distributions with implicit models (Huszár, 2017; Mohamed & Lakshminarayanan, 2016; Tran et al., 2017; Li & Turner, 2017; Mescheder et al., 2017; Shi et al., 2017). A typical example is transforming random noise via a deep neural network, leading to a non-invertible highly nonlinear mapping and hence an implicit distribution.

While an implicit variational distribution can be made highly flexible, it becomes necessary in each iteration to address the problem of density ratio estimation, which is often transformed into a problem related to learning generative adversarial networks (Goodfellow et al., 2014). In particular, a binary classifier, whose class probability is used for density ratio estimation, is trained in each iteration to discriminate the samples generated by the model from those by the variational distribution (Mohamed & Lakshminarayanan, 2016; Uehara et al., 2016; Mescheder et al., 2017). Controlling the bias and variance in density ratio estimation, however, is in general a very difficult problem, especially in high-dimensional settings (Sugiyama et al., 2012).

SIVI is related to the hierarchical variational model (HVM) (Ranganath et al., 2016; Maaløe et al., 2016) in having a hierarchical variational distribution, but there are two major distinctions: *1)* the HVM restricts the mixing distribution in the hierarchy to have an explicit PDF, which can be constructed with a Markov chain (Salimans et al., 2015), a mixture model (Ranganath et al., 2016), or a normalizing flow (Ranganath et al., 2016; Louizos & Welling, 2017) but cannot come from an implicit model. By contrast, SIVI requires the conditional distribution $q(\boldsymbol{z}\,|\,\boldsymbol{\psi})$ to have an explicit PDF, but does not impose such a constraint on the mixing distribution $q(\boldsymbol{\psi})$. In fact, any off-the-shelf reparameterizable implicit/explicit distribution can be used in SIVI, leading to considerably flexible $h_{\boldsymbol{\phi}}(\boldsymbol{z}) = \mathbb{E}_{\boldsymbol{\psi}\sim q_{\boldsymbol{\phi}}(\boldsymbol{\psi})} q(\boldsymbol{z}\,|\,\boldsymbol{\psi})$. Moreover, SIVI does not require $q(\boldsymbol{z}\,|\,\boldsymbol{\psi})$ to be reparameterizable for conditionally conjugate models. *2)* the HVM optimizes on a lower bound of the ELBO, constructed by adding a recursively estimated variational distribution that approximates $q(\boldsymbol{\psi}\,|\,\boldsymbol{z}) = q(\boldsymbol{z}\,|\,\boldsymbol{\psi})q(\boldsymbol{\psi})/h(\boldsymbol{z})$. By contrast, SIVI sandwiches the ELBO between two bounds, and directly optimizes on an asymptotically exact surrogate ELBO, which involves only simple-to-compute analytic density ratios.

# 5. Experiments

We implement SIVI in Tensorflow (Abadi et al., 2015) for a range of inference tasks. Note SIVI is a general purpose algorithm not relying on conjugacy, and has an inherent advantage over MCMC in being able to generate independent, and identically distributed ($iid$) posterior samples on the fly, this is, by feed-forward propagating $iid$ random noises through the inferred semi-implicit hierarchy. The toy examples show SIVI captures skewness, kurtosis, and multimodality. A negative binomial model shows SIVI can accurately capture the dependencies between latent variables. A bivariate count distribution example shows for a conditionally conjugate model, SIVI can utilize a non-reparameterizable variational distribution, without being plagued by the high variance of score function gradient estimation. With Bayesian logistic regression, we demonstrate that SIVI can either work alone as a black-box inference procedure for correlated latent variables, or directly expand MFVI by adding a mixing distribution, leading to accurate uncertainty estimation on par with that of MCMC. Last but not least, moving beyond the canonical Gaussian based variational autoencoder (VAE), SIVI helps construct semi-implicit VAE (SIVAE) to improve unsupervised feature learning and amortized inference.

## 5.1. Expressiveness of SIVI

We first show the expressiveness of SIVI by approximating various target distributions. As listed in Table 1, the conditional layer of SIVI is chosen to be as simple as an isotropic Gaussian (or log-normal) distribution $\mathcal{N}(\mathbf{0}, \sigma_0^2\mathbf{I})$. The implicit mixing layer is a multilayer perceptron (MLP), with layer widths $[30, 60, 30]$ and a ten dimensional isotropic Gaussian noise as its input. We fix $\sigma_0^2 = 0.1$ and optimize the implicit layer to minimize $\text{KL}(\mathbb{E}_{q_{\boldsymbol{\phi}}(\boldsymbol{\psi})} q(\boldsymbol{z}\,\vert\,\boldsymbol{\psi}) \vert\vert p(\boldsymbol{z}))$. As shown in Figure 1, despite having a fixed purposely misspecified explicit layer, by training a flexible implicit layer, the random samples from which are illustrated in Figure 8 of Appendix D, SIVI infers a sophisticated marginal variational distribution that accurately captures the skewness, kurtosis, and/or multimodality exhibited by the target one.

*Table 1.* Inference and target distributions

| $h(\boldsymbol{z}) = \mathbb{E}_{\boldsymbol{\psi}\sim q(\boldsymbol{\psi})} q(\boldsymbol{z}\,\vert\,\boldsymbol{\psi})$ | $p(\boldsymbol{z})$ |
|---|---|
| $z \sim \mathcal{N}(\psi, 0.1)$, $\psi \sim q(\psi)$ | Laplace$(z; \mu = 0, b = 2)$ |
| | $0.3\mathcal{N}(z; -2, 1) + 0.7\mathcal{N}(z; 2, 1)$ |
| $z \sim$ Log-Normal$(\psi, 0.1)$, $\psi \sim q(\psi)$ | Gamma$(z; 2, 1)$ |
| $\boldsymbol{z} \sim \mathcal{N}\left(\boldsymbol{\psi}, \begin{bmatrix} 0.1 & 0 \\ 0 & 0.1 \end{bmatrix}\right)$, $\boldsymbol{\psi} \sim q(\boldsymbol{\psi})$ | $0.5\mathcal{N}(\boldsymbol{z}; -2, I) + 0.5\mathcal{N}(\boldsymbol{z}; 2, I)$ |
| | $\mathcal{N}(z_1; z_2^2/4, 1)\mathcal{N}(z_2; 0, 4)$ |
| | $0.5\mathcal{N}\left(\boldsymbol{z}; 0, \begin{bmatrix} 2 & 1.8 \\ 1.8 & 2 \end{bmatrix}\right) + 0.5\mathcal{N}\left(\boldsymbol{z}; 0, \begin{bmatrix} 2 & -1.8 \\ -1.8 & 2 \end{bmatrix}\right)$ |

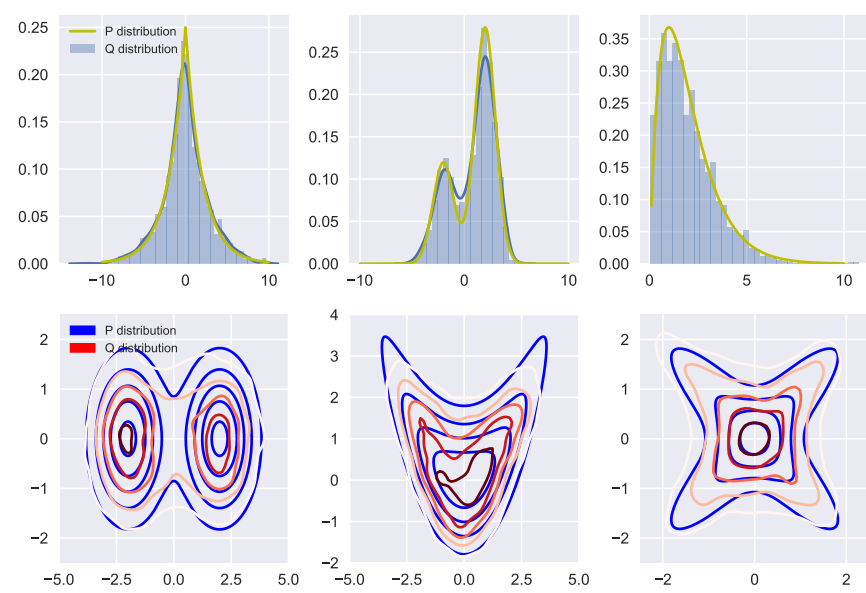

*Figure 1.* Approximating synthetic target distributions with SIVI

### 5.2. Negative Binomial Model

We consider a negative binomial (NB) distribution with the gamma and beta priors ($a = b = \alpha = \beta = 0.01$) as

$$x_i \overset{iid}{\sim} \text{NB}(r, p),\ r \sim \text{Gamma}(a, 1/b),\ p \sim \text{Beta}(\alpha, \beta),$$

for which the posterior $p(r, p\,\vert\,\{x_i\}_{1,N})$ is not tractable. In comparison to Gibbs sampling, it is shown in Zhou et al. (2012) that MFVI, which uses $q(r, p) = \text{Gamma}(r; \tilde{a}, 1/\tilde{b})\text{Beta}(p; \tilde{\alpha}, \tilde{\beta})$ to approximate the posterior, notably underestimates the variance. This caveat of MFVI motivates a semi-implicit variational distribution as

$$q(r, p\,\vert\,\boldsymbol{\psi}) = \text{Log-Normal}(r; \mu_r, \sigma_0^2)\text{Logit-Normal}(p; \mu_p, \sigma_0^2),$$
$$\boldsymbol{\psi} = (\mu_r, \mu_p) \sim q(\boldsymbol{\psi}), \sigma_0 = 0.1$$

where and an MLP based implicit $q(\boldsymbol{\psi})$, as in Section 5.1, is used by SIVI to capture the dependency between $r$ and $p$.

We apply Gibbs sampling, MFVI, and SIVI to a real overdispersed count dataset of Bliss & Fisher (1953) that records the number of adult red mites on each of the 150 randomly selected apple leaves. With $K = 1000$, as shown in Figure 2, SIVI improves MFVI in closely matching the posterior inferred by Gibbs sampling. Moreover, the mixing distribution $q(\boldsymbol{\psi})$ helps well capture the negative correlation between $r$ and $p$, as totally ignored by MFVI. The two-sample Kolmogorov-Smirnov (KS) distances, between 2000 posterior samples generated by SIVI and 2000 MCMC ones, are 0.0185 ($p$-value $= 0.88$) and 0.0200 ($p$-value $= 0.81$) for $r$ and $p$, respectively. By contrast, for MFVI and MCMC, they are 0.2695 ($p$-value $= 5.26 \times 10^{-64}$) and 0.2965 ($p$-value $= 2.21 \times 10^{-77}$), which significantly reject the null hypothesis that the posterior inferred by MFVI matches that by MCMC. How the performance is related to $K$ is shown in Figure 3 and Figures 9-10 of Appendix D, which suggests $K = 20$ achieves a nice compromise between complexity and accuracy, and as $K$ increases, the posterior inferred by SIVI quickly approaches that inferred by MCMC.

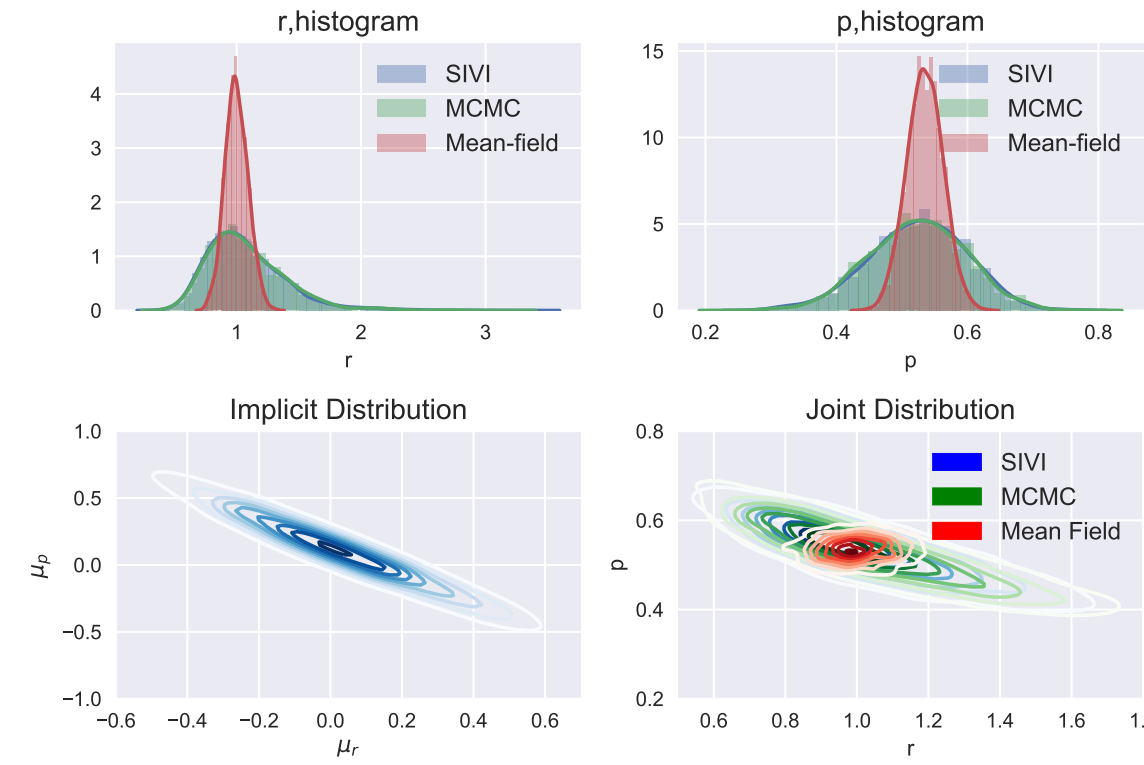

*Figure 2.* Top row: the marginal posteriors of $r$ and $p$ inferred by MFVI, SIVI, and MCMC. Bottom row: the inferred implicit mixing distribution $q(\boldsymbol{\psi})$ and joint posterior of $r$ and $p$.

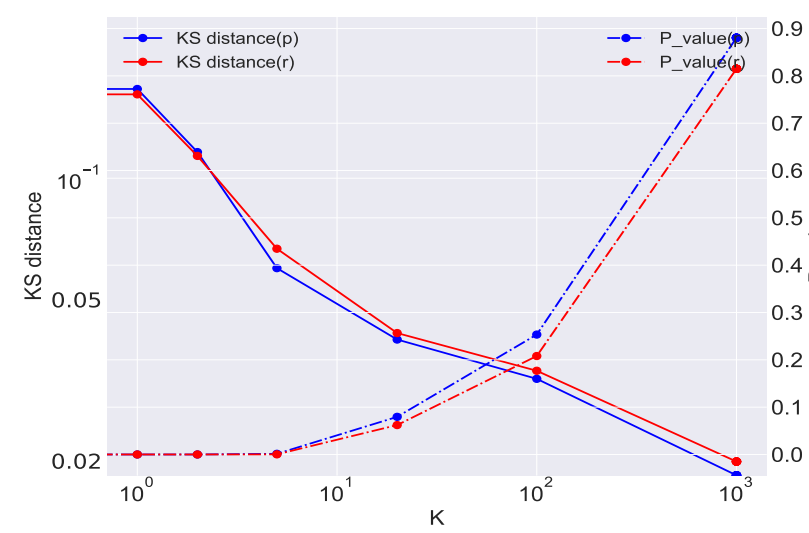

*Figure 3.* Kolmogorov-Smirnov (KS) distance and its corresponding $p$-value between the marginal posteriors of $r$ and $p$ inferred by SIVI and MCMC. SIVI rapidly improves as $K$ increases. See Appendix D for the corresponding plots of marginal posteriors.

### 5.3. Non-reparameterizable Variational Distribution

To show that SIVI can use a non-reparameterizable $q(\boldsymbol{z}\,\vert\,\boldsymbol{\psi})$ in a conditionally conjugate model, as discussed in Section 3.4, we apply it to infer the two parameters of the Poisson-logarithmic bivariate count distribution as $p(n_i, l_i\,\vert\,r, p) = r^{l_i} p^{n_i} (1-p)^r / Z_i$, where $l_i \in \{0, \ldots, n_i\}$ and $Z_i$ are normalization constants not related to $r > 0$ and $p \in (0, 1)$ (Zhou & Carin, 2015; Zhou et al., 2016). With $r \sim \text{Gamma}(a, 1/b)$ and $p \sim \text{Beta}(\alpha, \beta)$, while the joint posterior $p(r, p\,\vert\,\{n_i, l_i\}_{1,N})$ is intractable, the conditional posteriors of $r$ and $p$ follow the gamma and beta distributions, respectively. Although neither the gamma nor beta distribution is reparameterizable, SIVI multiplies them to construct a semi-implicit variational distribution as

$$q(r, p \mid \boldsymbol{\psi}) = \text{Gamma}(r; \psi_1, \psi_2)\text{Beta}(p; \psi_3, \psi_4),$$

where $\boldsymbol{\psi} = (\psi_1, \psi_2, \psi_3, \psi_4) \sim q(\boldsymbol{\psi})$ is similarly constructed as in Section 5.1. We set $K = 200$ for SIVI.

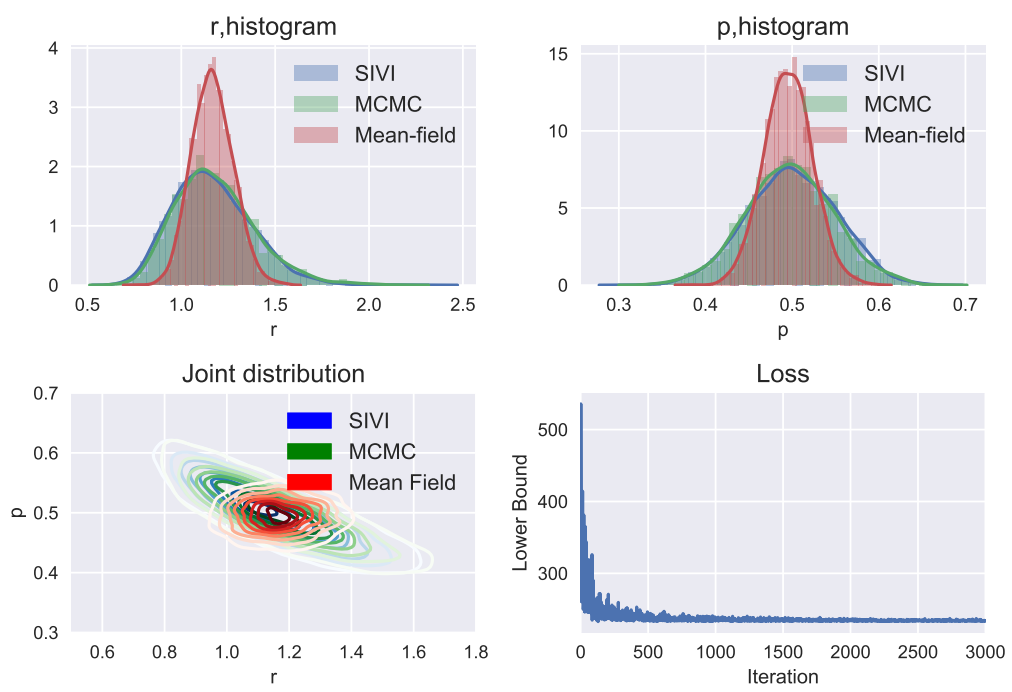

*Figure 4.* Top row: the marginal posteriors of $r$ and $p$ inferred by MFVI, SIVI, and MCMC. Bottom row: joint posteriors and the trance plots of $-\underline{\mathcal{L}}_{K_t}$ (subject to the difference of a constant).

As shown in Figure 4, despite the need of score function gradient estimation that is notorious for variance control, by utilizing conjugacy as in (11), SIVI well controls the variance of its gradient estimation, achieving accurate posterior estimation without requiring $q(\boldsymbol{z} \mid \boldsymbol{\psi})$ to be reparameterizable. By contrast, MFVI, which uses only the first summation term in (11) for gradient estimation, ignores covariance structure and notably underestimates posterior uncertainty.

### 5.4. Bayesian Logistic Regression

We compare SIVI with MFVI, Stein variational gradient descent (SVGD) of Liu & Wang (2016), and MCMC on Bayesian logistic regression, expressed as

$$y_i \sim \text{Bernoulli}[(1 + e^{-\boldsymbol{x}_i'\boldsymbol{\beta}})^{-1}], \;\; \boldsymbol{\beta} \sim \mathcal{N}(\mathbf{0}, \alpha^{-1}\mathbf{I}_{V+1}),$$

where $\boldsymbol{x}_i = (1, x_{i1}, \ldots, x_{iV})'$ are covariates, $y_i \in \{0, 1\}$ are binary response variables, and $\alpha$ is set as $0.01$. With the Pólya-Gamma data augmentation of Polson et al. (2013), we collect posterior MCMC samples of $\boldsymbol{\beta}$ using Gibbs sampling. For MFVI, the variational distribution is chosen as a multivariate normal (MVN) $\mathcal{N}(\boldsymbol{\beta}; \boldsymbol{\mu}, \boldsymbol{\Sigma})$, with a diagonal or full covariance matrix. For SIVI, we treat $\boldsymbol{\Sigma}$, diagonal or full, as a variational parameter, mix $\boldsymbol{\mu}$ with an MLP based implicit distribution, and set $K = 500$. We consider three datasets: *waveform*, *spam*, and *nodal*. The details on datasets and inference are deferred to Appendix B. On *waveform*, the algorithm converges in about 500 iterations, which takes about 40 seconds on a 2.4 GHz CPU. Note the results of SIVI with $K = 100$ (or 50), which takes about 12 (or 8) seconds for 500 iterations, are almost identical to those shown in Figures 5-8 with $K = 500$. Given the posterior captured by the semi-implicit hierarchy, SIVI takes 0.92 seconds to generate $50,000$ *iid* 22-dimensional $\boldsymbol{\beta}$'s.

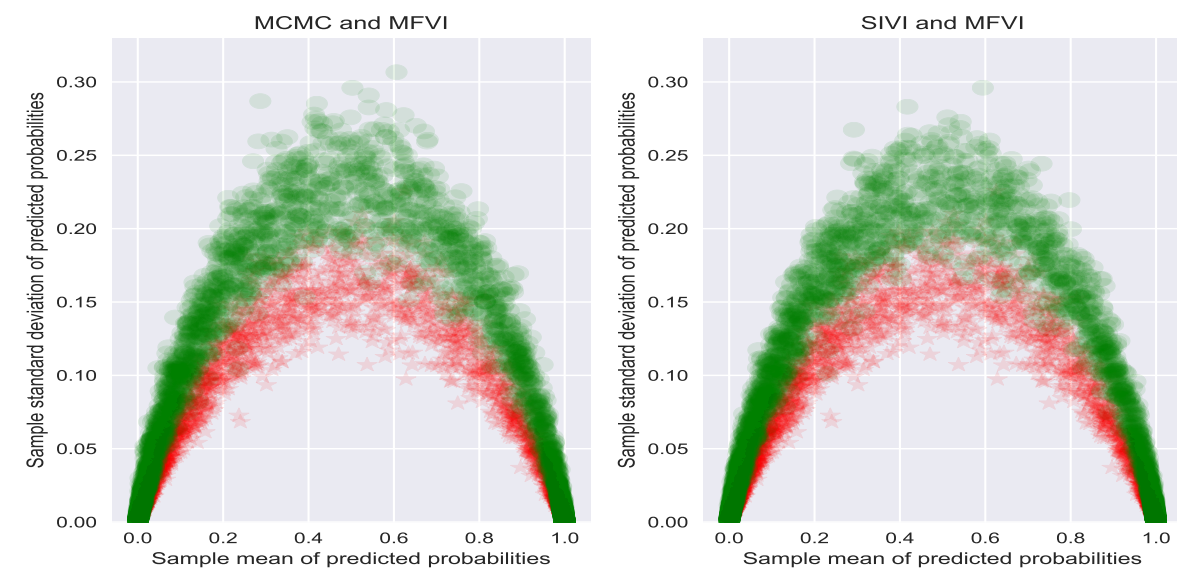

*Figure 5.* Comparison of MFVI (red) with a full covariance matrix, MCMC (green on left), and SIVI (green on right) with a full covariance matrix on quantifying predictive uncertainty for Bayesian logistic regression on *waveform*.

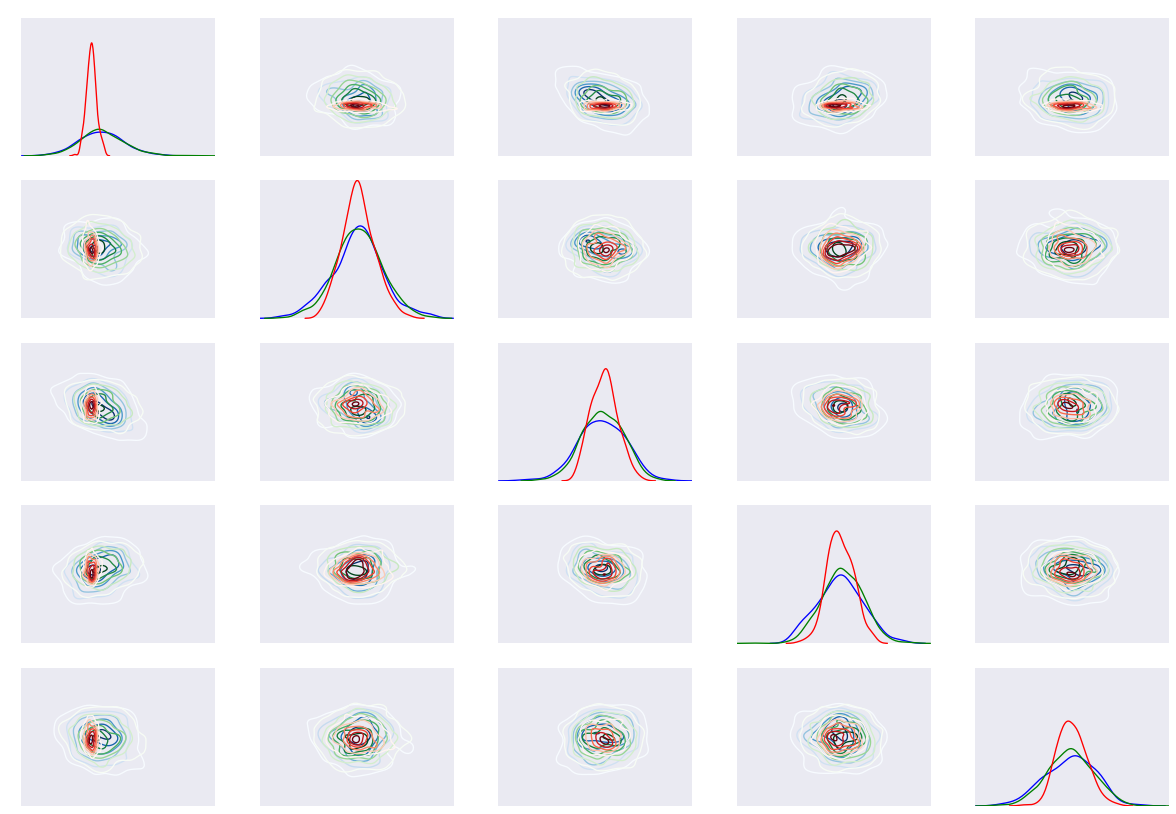

*Figure 6.* Marginal and pairwise joint posteriors for $(\beta_0, \ldots, \beta_4)$ inferred by MFVI (red, full covariance matrix), MCMC (blue), and SIVI (green, full covariance matrix) on *waveform*.

We collect $\boldsymbol{\beta}_j$ for $j = 1, \ldots, 1000$ to represent the inferred posterior $p(\boldsymbol{\beta} \mid \{\boldsymbol{x}_i, y_i\}_{1,N})$. For each test data $\boldsymbol{x}_{N+i}$, we calculate the predictive probabilities $1/(1 + e^{-\boldsymbol{x}_{N+i}^T \boldsymbol{\beta}_j})$ for all $j$ and compute its sample mean, and sample standard deviation that measures the uncertainty of the predictive distribution $p(y_{N+i} = 1 \mid \boldsymbol{x}_{N+i}, \{\boldsymbol{x}_i, y_i\}_{1,N})$. As shown in Figure 5, even with a full covariance matrix, the MVN variational distribution inferred by MFVI underestimates the uncertainty in out-of-sample prediction, let alone with a diagonal one, whereas SIVI, mixing the MVN with an MLP based implicit distribution, closely matches MCMC in uncertainty estimation. As shown in Figure 6, the underestimation of predictive uncertainty by MFVI can be attributed to variance underestimation for both univariate marginal and pairwise joint posteriors, which are, by contrast, well agreed on between SIVI and MCMC. Further examining the correlation coefficients of $\boldsymbol{\beta}$, shown in Figure 7, all the univariate marginals, shown in Figure 11 of Appendix D, and additional results, show in Figures 12-17 of Appendix D, it is revealed that SIVI well characterizes the posterior distribution of $\boldsymbol{\beta}$ and is only slightly negatively affected if its explicit layer is restricted with a diagonal covariance matrix, whereas MFVI with a diagonal/full covariance matrix and SVGD all misrepresent uncertainty. Note we have also tried modifying the code of variational boosting (Guo et al., 2016; Miller et al., 2017) for Bayesian logistic regression, but failed to obtain satisfactory results. We attribute the success of SIVI to its ability in better capturing the dependencies between $\beta_v$ and supporting a highly expressive non-Gaussian variational distribution by mixing an MVN with a flexible implicit distribution, whose parameters can be efficiently optimized via an asymptotically exact surrogate ELBO.

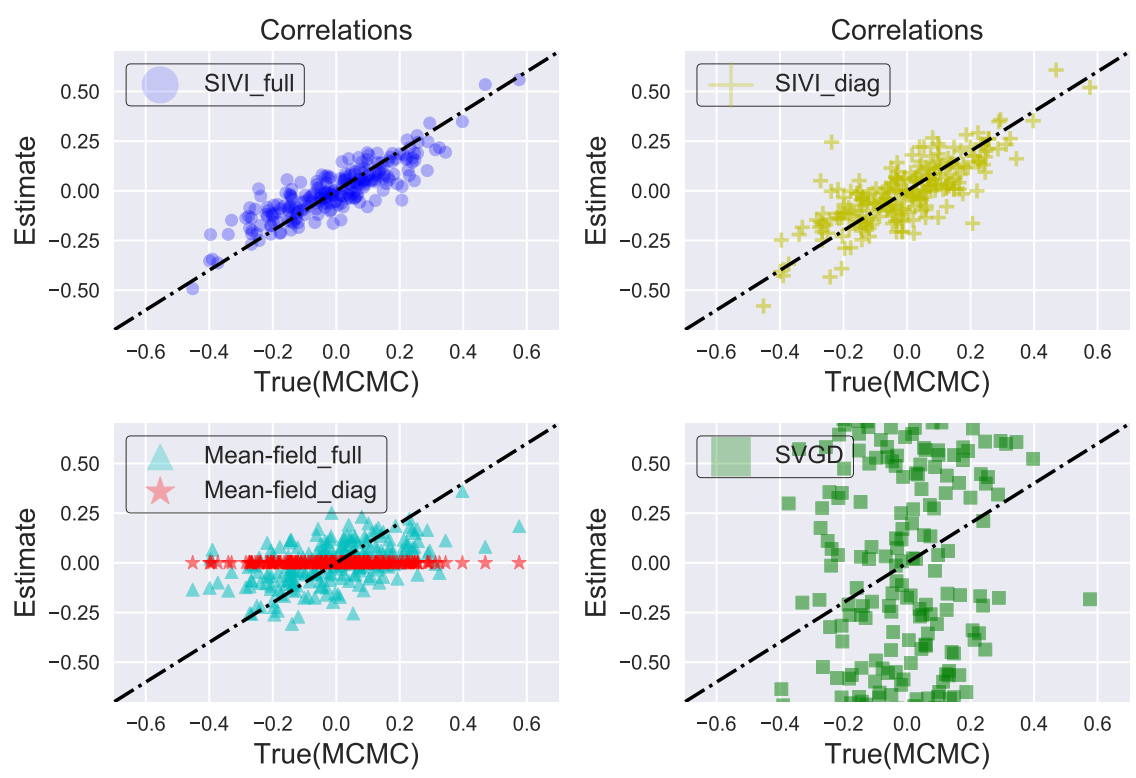

*Figure 7.* Comparing the correlation coefficients of $\boldsymbol{\beta}$ estimated from the posterior samples $\{\boldsymbol{\beta}_i\}_{i=1:1000}$ of SIVI with that of MCMC on *waveform* for SIVI with a full/diagonal covariance matrix, MFVI with a full/diagonal covariance matrix, and SVGD.

### 5.5. Semi-Implicit Variational Autoencoder

Variational Auto-encoder (VAE) (Kingma & Welling, 2013; Rezende et al., 2014), widely used for unsupervised feature learning and amortized inference, infers encoder parameter $\boldsymbol{\phi}$ and decoder parameter $\boldsymbol{\theta}$ to maximize the ELBO as

$$\mathcal{L}(\boldsymbol{\phi},\boldsymbol{\theta}) = \mathbb{E}_{\boldsymbol{z}\sim q_{\boldsymbol{\phi}}(\boldsymbol{z}\,|\,\boldsymbol{x})}[\log(p_{\boldsymbol{\theta}}(\boldsymbol{x}\,|\,\boldsymbol{z}))] - \mathrm{KL}(q_{\boldsymbol{\phi}}(\boldsymbol{z}\,|\,\boldsymbol{x})||p(\boldsymbol{z})).$$

The encoder distribution $q_{\boldsymbol{\phi}}(\boldsymbol{z}\,|\,\boldsymbol{x})$ is required to be reparameterizable and analytically evaluable, which usually restricts it to a small exponential family. In particular, a canonical encoder is $q_{\boldsymbol{\phi}}(\boldsymbol{z}\,|\,\boldsymbol{x}) = \mathcal{N}(\boldsymbol{z}\,|\,\boldsymbol{\mu}(\boldsymbol{x},\boldsymbol{\phi}),\boldsymbol{\Sigma}(\boldsymbol{x},\boldsymbol{\phi}))$, where the Gaussian parameters are deterministically transformed from the observations $\boldsymbol{x}$, via non-probabilistic deep neural networks parameterized by $\boldsymbol{\phi}$. Thus, given observation $\boldsymbol{x}_i$, its corresponding code $\boldsymbol{z}_i$ is forced to follow a Gaussian distribution, no matter how powerful the deep neural networks are. The Gaussian assumption, however, is often too restrictive to model skewness, kurtosis, and multimodality.

To this end, rather than using a single-stochastic-layer encoder, we use SIVI that can add as many stochastic layers as needed, as long as the first stochastic layer $q_{\boldsymbol{\phi}}(\boldsymbol{z}\,|\,\boldsymbol{x})$ is reparameterizable and has an analytic PDF, and the layers added after are reparameterizable and simple to sample from. More specifically, we construct semi-implicit VAE (SIVAE) by using a hierarchical encoder that injects random noise at $M$ different stochastic layers as

$$\begin{aligned}
&\boldsymbol{\ell}_t = T_t(\boldsymbol{\ell}_{t-1},\boldsymbol{\epsilon}_t,\boldsymbol{x};\boldsymbol{\phi}),\ \boldsymbol{\epsilon}_t \sim q_t(\boldsymbol{\epsilon}),\ t=1,\dots,M,\\
&\boldsymbol{\mu}(\boldsymbol{x},\boldsymbol{\phi}) = f(\boldsymbol{\ell}_M,\boldsymbol{x};\boldsymbol{\phi}),\ \ \boldsymbol{\Sigma}(\boldsymbol{x},\boldsymbol{\phi}) = g(\boldsymbol{\ell}_M,\boldsymbol{x};\boldsymbol{\phi}),\\
&q_{\boldsymbol{\phi}}(\boldsymbol{z}\,|\,\boldsymbol{x},\boldsymbol{\mu},\boldsymbol{\Sigma}) = \mathcal{N}(\boldsymbol{\mu}(\boldsymbol{x},\boldsymbol{\phi}),\boldsymbol{\Sigma}(\boldsymbol{x},\boldsymbol{\phi})),
\end{aligned} \tag{12}$$

where $\boldsymbol{\ell}_0 = \emptyset$ and $T_t$, $f$, and $g$ are all deterministic neural networks. Note given data $\boldsymbol{x}_i$, $\boldsymbol{\mu}(\boldsymbol{x}_i,\boldsymbol{\phi})$, $\boldsymbol{\Sigma}(\boldsymbol{x}_i,\boldsymbol{\phi})$ are now random variables rather than following vanilla VAE to assume deterministic values. This moves the encoder variational distribution beyond a simple Gaussian form.

To benchmark the performance of SIVAE, we consider the MNIST dataset that is stochastically binarized as in Salakhutdinov & Murray (2008). We use 55,000 for training and use the 10,000 observations in the testing set for performance evaluation. Similar to existing VAEs, we choose Bernoulli units, linked to a fully-connected neural network with two 500-unit hidden layers, as the decoder. Distinct from existing VAEs, whose encoders are often restricted to have a single stochastic layer, SIVI allows SIVAE to use an MVN as its first stochastic layer, and draw the parameters of the MVN from $M = 3$ stochastic layers, whose structure is described in detail in Appendix C. As shown in Table 2 of Appendix C, SIVAE achieves a negative log evidence of $84.07$, which is further reduced to $83.25$ if choosing importance reweighing with $\tilde{K} = 10$. In comparison to other VAEs with a comparable single-stochastic-layer decoder, SIVAE achieves state-of-the-art performance by mixing an MVN with an implicit distribution defined as in (12) to construct a flexible encoder, whose marginal variational distribution is no longer restricted to the MVN distribution. We leave it for future study on further improving SIVAE by replacing the encoder MVN explicit layer with a normalizing flow, and adding convolution/autoregression to enrich the encoder's implicit distribution and/or the decoder.

## 6. Conclusions

Combining the advantages of having analytic point-wise evaluable density ratios and tractable computation via Monte Carlo estimation, semi-implicit variational inference (SIVI) is proposed either as a black-box inference procedure, or to enrich mean-field variational inference with a flexible (implicit) mixing distribution. By designing a surrogate evidence lower bound that is asymptotically exact, SIVI establishes an optimization problem amenable to gradient ascent, without compromising the expressiveness of its semi-implicit variational distribution. Flexible but simple to optimize, SIVI approaches the accuracy of MCMC in quantifying posterior uncertainty in a wide variety of inference tasks, and is not constrained by conjugacy, often runs faster, and can generate *iid* posterior samples on the fly via the inferred stochastic variational inference network.

## Acknowledgements

The authors thank Texas Advanced Computing Center (TACC) for computational support.

# Semi-Implicit Variational Inference: Supplementary Material

**Mingzhang Yin and Mingyuan Zhou**

**Algorithm 1** Semi-Implicit Variational Inference (SIVI)

**input** : Data $\{x_i\}_{1:N}$, joint likelihood $p(\boldsymbol{x}, \boldsymbol{z})$, explicit variational distribution $q_{\boldsymbol{\xi}}(\boldsymbol{z} \mid \boldsymbol{\psi})$ with reparameterization $\boldsymbol{z} = f(\boldsymbol{\epsilon}, \boldsymbol{\xi}, \boldsymbol{\psi})$, $\boldsymbol{\epsilon} \sim p(\boldsymbol{\epsilon})$, implicit layer neural network $T_{\boldsymbol{\phi}}(\epsilon)$ and source of randomness $q(\boldsymbol{\epsilon})$

**output** : Variational parameter $\boldsymbol{\xi}$ for the conditional distribution $q_{\boldsymbol{\xi}}(\boldsymbol{z} \mid \boldsymbol{\psi})$, variational parameter $\boldsymbol{\phi}$ for the mixing distribution $q_{\boldsymbol{\phi}}(\boldsymbol{\psi})$

Initialize $\boldsymbol{\xi}$ and $\boldsymbol{\phi}$ randomly

**while** *not converged* **do**

  Set $\underline{L}_{K_t} = 0$, $\rho_t$ and $\eta_t$ as step sizes, and $K_t \geq 0$ as a non-decreasing integer; Sample $\boldsymbol{\psi}^{(k)} = T_{\boldsymbol{\phi}}(\boldsymbol{\epsilon}^{(k)})$, $\boldsymbol{\epsilon}^{(k)} \sim q(\boldsymbol{\epsilon})$ for $k = 1, \dots, K_t$; take subsample $\mathbf{x} = \{x_i\}_{i_1:i_M}$

  **for** $j = 1$ **to** $J$ **do**

    Sample $\boldsymbol{\psi}_j = T_{\boldsymbol{\phi}}(\boldsymbol{\epsilon}_j)$, $\boldsymbol{\epsilon}_j \sim q(\boldsymbol{\epsilon})$

    Sample $\boldsymbol{z}_j = f(\tilde{\boldsymbol{\epsilon}}_j, \boldsymbol{\xi}, \boldsymbol{\psi}_j)$, $\tilde{\boldsymbol{\epsilon}}_j \sim p(\boldsymbol{\varepsilon})$

    $\underline{L}_{K_t} = \underline{L}_{K_t} + \frac{1}{J}\big\{ - \log \frac{1}{K_t+1}\big[\sum_{k=1}^{K_t} q_{\boldsymbol{\xi}}(\boldsymbol{z}_j \mid \boldsymbol{\psi}^{(k)}) + q_{\boldsymbol{\xi}}(\boldsymbol{z}_j \mid \boldsymbol{\psi}_j)\big] + \frac{N}{M}\log p(\mathbf{x} \mid \boldsymbol{z}_j) + \log p(\boldsymbol{z}_j)\big\}$

  **end**

  $t = t + 1$

  $\boldsymbol{\xi} = \boldsymbol{\xi} + \rho_t \nabla_{\boldsymbol{\xi}} \underline{L}_{K_t}\big(\{\boldsymbol{\psi}^{(k)}\}_{1,K_t}, \{\boldsymbol{\psi}_j\}_{1,J}, \{\boldsymbol{z}_j\}_{1,J}\big)$

  $\boldsymbol{\phi} = \boldsymbol{\phi} + \eta_t \nabla_{\boldsymbol{\phi}} \underline{L}_{K_t}\big(\{\boldsymbol{\psi}^{(k)}\}_{1,K_t}, \{\boldsymbol{\psi}_j\}_{1,J}, \{\boldsymbol{z}_j\}_{1,J}\big)$

**end**

## A. Proofs

*Proof of Inequility* (3). To prove a functional form of Jensen's Inequality, let $h(\boldsymbol{z}) = \mathbb{E}_{\boldsymbol{\psi} \sim q_{\boldsymbol{\phi}}(\boldsymbol{\psi})} q(\boldsymbol{z}|\boldsymbol{\psi})$ and $\langle f, g \rangle_{L^2} = \int f(\boldsymbol{z}) g(\boldsymbol{z}) d\boldsymbol{z}$. From Theorem 1, we have convexity, and according to Theorem 6.2.1. of Kurdila & Zabarankin (2005), we have an equivalent first-order definition for convexity as

$$\mathrm{KL}(q(\boldsymbol{z}|\boldsymbol{\psi}) || p(\boldsymbol{z})) \geq \mathrm{KL}(h(\boldsymbol{z}) || p) + \langle q(\boldsymbol{z}|\boldsymbol{\psi}) - h(\boldsymbol{z}), \nabla_q \mathrm{KL}(q||p)|_{h(\boldsymbol{z})} \rangle_{L^2}$$

Taking the expectation with respect to $\boldsymbol{\psi} \sim q_{\boldsymbol{\phi}}(\boldsymbol{\psi})$ on both sides, we have

$$\begin{aligned}
&\mathbb{E}_{\boldsymbol{\psi} \sim q_{\boldsymbol{\phi}}(\boldsymbol{\psi})} \mathrm{KL}(q(\boldsymbol{z}|\boldsymbol{\psi}) || p(\boldsymbol{z})) \\
&\geq \mathrm{KL}(h(\boldsymbol{z}) || p(\boldsymbol{z})) \\
&\quad + \mathbb{E}_{\boldsymbol{\psi} \sim q_{\boldsymbol{\phi}}(\boldsymbol{\psi})} [\langle q(\boldsymbol{z}|\boldsymbol{\psi}) - h(\boldsymbol{z}), \nabla_q \mathrm{KL}(q||p)|_{h(\boldsymbol{z})} \rangle_{L^2}] \\
&= \mathrm{KL}(h(\boldsymbol{z}) || p(\boldsymbol{z})) \\
&= \mathrm{KL}(\mathbb{E}_{\boldsymbol{\psi} \sim q_{\boldsymbol{\phi}}(\boldsymbol{\psi})} q(\boldsymbol{z}|\boldsymbol{\psi}) || p(\boldsymbol{z})).
\end{aligned}$$

□

*Proof of Proposition 1.* We show that directly maximizing the lower bound $\underline{\mathcal{L}}$ of ELBO in (4) may drive $q(\boldsymbol{\psi})$ towards degeneracy. For VI that uses $q(\boldsymbol{z} \mid \boldsymbol{\psi})$ as its variational distribution, if supposing $\boldsymbol{\psi}^*$ is the optimum variational parameter, which means

$$\boldsymbol{\psi}^* = \arg\max_{\boldsymbol{\psi}} -\mathbb{E}_{\boldsymbol{z} \sim q(\boldsymbol{z}|\boldsymbol{\psi})} \log \frac{q(\boldsymbol{z}|\boldsymbol{\psi})}{p(\boldsymbol{x}, \boldsymbol{z})},$$

then we have

$$\begin{aligned}
\underline{\mathcal{L}} &= -\mathbb{E}_{\boldsymbol{\psi} \sim q_{\boldsymbol{\phi}}(\boldsymbol{\psi})} \mathbb{E}_{\boldsymbol{z} \sim q(\boldsymbol{z}|\boldsymbol{\psi})} \log \frac{q(\boldsymbol{z}|\boldsymbol{\psi})}{p(\boldsymbol{x}, \boldsymbol{z})} \\
&= \int q_{\boldsymbol{\phi}}(\boldsymbol{\psi}) [-\mathbb{E}_{\boldsymbol{z} \sim q(\boldsymbol{z}|\boldsymbol{\psi})} \log \frac{q(\boldsymbol{z}|\boldsymbol{\psi})}{p(\boldsymbol{x}, \boldsymbol{z})}] d\boldsymbol{\psi} \\
&\leq \int q_{\boldsymbol{\phi}}(\boldsymbol{\psi}) d\boldsymbol{\psi} [-\mathbb{E}_{\boldsymbol{z} \sim q(\boldsymbol{z}|\boldsymbol{\psi}^*)} \log \frac{q(\boldsymbol{z}|\boldsymbol{\psi}^*)}{p(\boldsymbol{x}, \boldsymbol{z})}] \\
&= -\mathbb{E}_{\boldsymbol{z} \sim q(\boldsymbol{z}|\boldsymbol{\psi}^*)} \log \frac{q(\boldsymbol{z}|\boldsymbol{\psi}^*)}{p(\boldsymbol{x}, \boldsymbol{z})}.
\end{aligned}$$

The equality in the above equation is reached if and only if $q(\boldsymbol{\psi}) = \delta_{\boldsymbol{\psi}^*}(\boldsymbol{\psi})$, which means the mixing distribution degenerates to a point mass density and hence SIVI degenerates to vanilla VI. □

*Proof of Proposition 2.* $B_0 = 0$ is trivial. Denote $\boldsymbol{\psi}^{(0)} = \psi_v$. For $iid$ samples $\boldsymbol{\psi}^{(k)} \sim q_{\boldsymbol{\phi}}(\boldsymbol{\psi})$, when $K \to \infty$, by the strong law of large numbers, $\tilde{h}_K(\boldsymbol{z}) = \frac{\sum_{k=0}^{K} q(\boldsymbol{z} \mid \boldsymbol{\psi}^{(k)})}{K+1}$ converges almost surely to $\mathbb{E}_{q_{\boldsymbol{\phi}}(\boldsymbol{\psi})} q(\boldsymbol{z} \mid \boldsymbol{\psi}) = h_{\boldsymbol{\phi}}(\boldsymbol{z})$. To prove (6), by the strong law of large numbers, we first rewrite it as the limit of a double sequence $S(K, J)$, where $K, J \in \{1, 2, \dots, \}$, and check the condition for the interchange of iterated limits (Rudin, 1964; Habil, 2016): i) The double limit exists; ii) Fixing one index of the double sequence, the one side limit exists for the other index .

$$\begin{aligned}
&\lim_{K \to \infty} \mathbb{E}_{\boldsymbol{\psi}^{(0)}, \boldsymbol{\psi}^{(1)}, \cdots, \boldsymbol{\psi}^{(K)} \sim q(\boldsymbol{\psi})} \log \frac{\sum_{k=0}^{K} q(\boldsymbol{z} \mid \boldsymbol{\psi}^{(k)})}{K+1} \\
&= \lim_{K \to \infty} \lim_{J \to \infty} \frac{1}{J} \sum_{j=1}^{J} \log \frac{1}{K+1} \sum_{k=0}^{K} q(\boldsymbol{z} \mid \boldsymbol{\psi}_j^{(k)}) \\
&\triangleq \lim_{K \to \infty} \lim_{J \to \infty} S(K, J).
\end{aligned}$$

Here $\boldsymbol{\psi}_j^{(k)}$ are $iid$ samples from $q(\boldsymbol{\psi})$. For i) we show double limit $\lim_{K,J \to \infty} S(K, J) = \log h(\boldsymbol{z})$. For $\forall \epsilon > 0$, $\exists N(\epsilon)$, when $K, J > N(\epsilon)$, $|\log \frac{1}{K+1} \sum_{k=0}^{K} q(\boldsymbol{z} \mid \boldsymbol{\psi}_j^{(k)}) - \log h(\boldsymbol{z})| < \epsilon$ thanks to the law of large numbers, then

$$\begin{aligned}
&\left| \sum_{j=1}^{J} \log \frac{1}{K+1} \sum_{k=0}^{K} q(\boldsymbol{z} \mid \boldsymbol{\psi}_j^{(k)}) - J \log h(\boldsymbol{z}) \right| \\
&\leq \sum_{j=1}^{J} \left| \log \frac{1}{K+1} \sum_{k=0}^{K} q(\boldsymbol{z} \mid \boldsymbol{\psi}_j^{(k)}) - \log h(\boldsymbol{z}) \right| \leq J\epsilon.
\end{aligned}$$

Deviding both sides by $J$ we get $|S(K,J) - \log h(\boldsymbol{z})| \leq \epsilon$ when $K, J > N(\epsilon)$. By definition, we have $\lim_{K,J\to\infty} S(K,J) = \log h(\boldsymbol{z})$.
ii) for each fixed $J \in \mathbb{N}$, $\lim_{K\to\infty} S(K,J) = \log h(\boldsymbol{z})$ exists; for each fixed $K \in \mathbb{N}$, $\lim_{J\to\infty} S(K,J) = \mathbb{E}_{\boldsymbol{\psi}^{(0)},\boldsymbol{\psi}^{(1)},\cdots,\boldsymbol{\psi}^{(K)}\sim q(\boldsymbol{\psi})} \log \frac{\sum_{k=0}^K q(\boldsymbol{z}\,|\,\boldsymbol{\psi}^{(k)})}{K+1} \leq \log h(\boldsymbol{z})$ also exists. The limitation can then be interchanged and (6) is proved. Therefore, we have

$$
\begin{aligned}
&\lim_{k\to\infty} \underline{\mathcal{L}}_k = \underline{\mathcal{L}} + \mathbb{E}_{\boldsymbol{\psi}}\mathrm{KL}(q(\boldsymbol{z}\,|\,\boldsymbol{\psi})||h_{\boldsymbol{\phi}}(\boldsymbol{z}))\\
&= \mathbb{E}_{\boldsymbol{\psi}\sim q(\boldsymbol{\psi})}\mathbb{E}_{\boldsymbol{z}\sim q(\boldsymbol{z}\,|\,\boldsymbol{\psi})}\left[\log\frac{q(\boldsymbol{z}\,|\,\boldsymbol{\psi})}{h_{\boldsymbol{\phi}}(\boldsymbol{z})} - \log\frac{q(\boldsymbol{z}\,|\,\boldsymbol{\psi})}{p(x,\boldsymbol{z})}\right]\\
&= -\mathbb{E}_{\boldsymbol{\psi}\sim q(\boldsymbol{\psi})}\mathbb{E}_{\boldsymbol{z}\sim q(\boldsymbol{z}\,|\,\boldsymbol{\psi})}\log\frac{h_{\boldsymbol{\phi}}(\boldsymbol{z})}{p(x,\boldsymbol{z})}\\
&= -\mathbb{E}_{\boldsymbol{z}\sim h_{\boldsymbol{\phi}}(\boldsymbol{z})}\log\frac{h_{\boldsymbol{\phi}}(\boldsymbol{z})}{p(x,\boldsymbol{z})} = \mathcal{L}.
\end{aligned}
$$

□

*Proof of Proposition 3.* Assume integer $K > M > 0$. Let $\mathcal{I}$ be the set that consists of all the subsets of $\{1,\cdots,K\}$ with cardinality $M$. Let $I$ be a discrete uniform random variable and for element $\{i_1,\cdots,i_M\}\in\mathcal{I}$, $P(I=\{i_1,\cdots,i_M\}) = \frac{1}{\binom{K}{M}}$. We have $\mathbb{E}_I\frac{1}{M}\sum_{i\in I} q(\boldsymbol{z}\,|\,\boldsymbol{\psi}^i) = \frac{1}{K}\sum_{i=1}^K q(\boldsymbol{z}\,|\,\boldsymbol{\psi}^i)$. To show $\bar{\mathcal{L}}_K = \bar{\mathcal{L}} - A_K$ is monotonic decreasing, we only need to show $A_K$ is monotonic increasing:

$$
\begin{aligned}
A_K =& \mathbb{E}_{\boldsymbol{\psi}\sim q(\boldsymbol{\psi})}\mathbb{E}_{\boldsymbol{z}\sim h_{\boldsymbol{\phi}}(\boldsymbol{z})}\mathbb{E}_{\boldsymbol{\psi}^{(1)},\cdots,\boldsymbol{\psi}^{(K)}\sim q(\boldsymbol{\psi})}\\
&\log\frac{\frac{1}{K}\sum_{i=1}^K q(\boldsymbol{z}\,|\,\boldsymbol{\psi}^{(i)})}{q(\boldsymbol{z}\,|\,\boldsymbol{\psi})}\\
=&\mathbb{E}_{\boldsymbol{\psi}\sim q(\boldsymbol{\psi})}\mathbb{E}_{\boldsymbol{z}\sim h_{\boldsymbol{\phi}}(\boldsymbol{z})}\mathbb{E}_{\boldsymbol{\psi}^{(1)},\cdots,\boldsymbol{\psi}^{(K)}\sim q(\boldsymbol{\psi})}\\
&\log\mathbb{E}_I\left[\frac{\frac{1}{M}\sum_{i\in I} q(\boldsymbol{z}\,|\,\boldsymbol{\psi}^{(i)})}{q(\boldsymbol{z}\,|\,\boldsymbol{\psi})}\right]\\
\geq&\mathbb{E}_{\boldsymbol{\psi}\sim q(\boldsymbol{\psi})}\mathbb{E}_{\boldsymbol{z}\sim h_{\boldsymbol{\phi}}(\boldsymbol{z})}\mathbb{E}_{\boldsymbol{\psi}^{(1)},\cdots,\boldsymbol{\psi}^{(K)}\sim q(\boldsymbol{\psi})}\\
&\mathbb{E}_I\log\frac{\frac{1}{M}\sum_{i\in I} q(\boldsymbol{z}\,|\,\boldsymbol{\psi}^{(i)})}{q(\boldsymbol{z}\,|\,\boldsymbol{\psi})}\\
=&\mathbb{E}_{\boldsymbol{\psi}\sim q(\boldsymbol{\psi})}\mathbb{E}_{\boldsymbol{z}\sim h_{\boldsymbol{\phi}}(\boldsymbol{z})}\mathbb{E}_{\boldsymbol{\psi}^{(1)},\cdots,\boldsymbol{\psi}^{(M)}\sim q(\boldsymbol{\psi})}\\
&\log\frac{\frac{1}{M}\sum_{i=1}^M q(\boldsymbol{z}\,|\,\boldsymbol{\psi}^{(i)})}{q(\boldsymbol{z}\,|\,\boldsymbol{\psi})}\\
=&A_M.
\end{aligned}
$$

We now show $\lim_{K\to\infty}\bar{\mathcal{L}}_K = \mathcal{L}$. Again, by the strong law of large numbers, $\frac{1}{K}\sum_{i=1}^K q(\boldsymbol{z}\,|\,\boldsymbol{\psi}^{(i)})$ converges almost surely to $\mathbb{E}_{\boldsymbol{\psi}\sim q_{\boldsymbol{\phi}}(\boldsymbol{\psi})} q(\boldsymbol{z}\,|\,\boldsymbol{\psi}) = h_{\boldsymbol{\phi}}(\boldsymbol{z})$ and hence

$$
\begin{aligned}
&\lim_{K\to\infty}\bar{\mathcal{L}}_K = \bar{\mathcal{L}} + \mathbb{E}_{\boldsymbol{\psi}}\mathrm{KL}(h_{\boldsymbol{\phi}}(\boldsymbol{z})||q(\boldsymbol{z}\,|\,\boldsymbol{\psi}))\\
&= -\mathbb{E}_{\boldsymbol{z}\sim h_{\boldsymbol{\phi}}(\boldsymbol{z})}\mathbb{E}_{\boldsymbol{\psi}\sim q(\boldsymbol{\psi})}\left[\log\frac{q(\boldsymbol{z}\,|\,\boldsymbol{\psi})}{p(x,\boldsymbol{z})} + \log\frac{h_{\boldsymbol{\phi}}(\boldsymbol{z})}{q(\boldsymbol{z}\,|\,\boldsymbol{\psi})}\right]\\
&= \mathcal{L}.
\end{aligned}
$$

□

*Proof of Equation (11).* The gradient of $B_K$ with respect to $\boldsymbol{\phi}$ can be expressed as

$$
\begin{aligned}
&\nabla_{\boldsymbol{\phi}} B_K = \nabla_{\boldsymbol{\phi}}\mathbb{E}_{\boldsymbol{\psi}\sim q_{\boldsymbol{\phi}}(\boldsymbol{\psi})}\mathbb{E}_{\boldsymbol{\psi}^{(1)},\ldots,\boldsymbol{\psi}^{(K)}\sim q_{\boldsymbol{\phi}}(\boldsymbol{\psi})}\Bigg[\\
&\quad\mathrm{KL}\left(q(\boldsymbol{z}\,|\,\boldsymbol{\psi})\middle|\middle|\frac{q(\boldsymbol{z}\,|\,\boldsymbol{\psi})+\sum_{k=1}^K q(\boldsymbol{z}\,|\,\boldsymbol{\psi}^{(k)})}{K+1}\right)\Bigg]\\
&= \mathbb{E}_{\boldsymbol{\epsilon},\boldsymbol{\epsilon}^{(1)},\ldots,\boldsymbol{\epsilon}^{(K)}\sim p(\boldsymbol{\epsilon})}\nabla_{\boldsymbol{\phi}}\mathbb{E}_{\boldsymbol{z}\sim q(\boldsymbol{z}\,|\,T_{\boldsymbol{\phi}}(\boldsymbol{\epsilon}))}\Bigg[\\
&\quad\log\frac{q(\boldsymbol{z}\,|\,T_{\boldsymbol{\phi}}(\boldsymbol{\epsilon}))}{\frac{q(\boldsymbol{z}\,|\,T_{\boldsymbol{\phi}}(\boldsymbol{\epsilon}))+\sum_{k=1}^K q(\boldsymbol{z}\,|\,T_{\boldsymbol{\phi}}(\boldsymbol{\epsilon}^{(k)}))}{K+1}}\Bigg]\\
&= \mathbb{E}_{\boldsymbol{\epsilon},\ldots,\boldsymbol{\epsilon}^{(K)}}\nabla_{\boldsymbol{\phi}}\mathbb{E}_{\boldsymbol{z}\sim q(\boldsymbol{z}\,|\,T_{\boldsymbol{\phi}}(\boldsymbol{\epsilon}))}\log\left[r_{\boldsymbol{\phi}}(\boldsymbol{z},\boldsymbol{\epsilon},\boldsymbol{\epsilon}^{(1:K)})\right]\\
&= \mathbb{E}_{\boldsymbol{\epsilon},\ldots,\boldsymbol{\epsilon}^{(K)}\sim p(\boldsymbol{\epsilon})}\mathbb{E}_{\boldsymbol{z}\sim q(\boldsymbol{z}\,|\,T_{\boldsymbol{\phi}}(\boldsymbol{\epsilon}))}\Big\{\\
&\quad q(\boldsymbol{z}\,|\,T_{\boldsymbol{\phi}}(\boldsymbol{\epsilon}))\nabla_{\boldsymbol{\phi}}\log\left[r_{\boldsymbol{\phi}}(\boldsymbol{z},\boldsymbol{\epsilon},\boldsymbol{\epsilon}^{(1:K)})\right]\\
&\quad+\left[\nabla_{\boldsymbol{\phi}}\log q(\boldsymbol{z}\,|\,T_{\boldsymbol{\phi}}(\boldsymbol{\epsilon}))\right]\log\left[r_{\boldsymbol{\phi}}(\boldsymbol{z},\boldsymbol{\epsilon},\boldsymbol{\epsilon}^{(1:K)})\right]\Big\}.
\end{aligned}
$$

□

## B. Bayesian Logistic Regression

We consider datesets *waveform* ($n = 5000$, $V = 21$, and $400/4600$ for training/testing), *spam* ($n = 3000$, $V = 2$, and 2000/1000 for training/testing), and *nodal* ($n = 53$, $V = 5$, and 25/28 for training/testing). The training-set-size to feature-dimension ratio $n_{\text{train}}/V$ varies in these three datasets, and we expect the posterior uncertainty to be large if this ratio is small.

The contribution of observation $i$ to the likelihood can be expressed as

$$
P(y_i\,|\,\boldsymbol{x}_i,\boldsymbol{\beta}) = \frac{e^{y\boldsymbol{x}_i'\boldsymbol{\beta}}}{1+e^{\boldsymbol{x}_i'\boldsymbol{\beta}}} \quad \propto e^{(y-\frac{1}{2})\boldsymbol{x}_i'\boldsymbol{\beta}}\mathbb{E}_{\omega_i}\left[e^{-\frac{\omega_i(\boldsymbol{x}_i'\boldsymbol{\beta})^2}{2}}\right],
$$

where the expectation is taken respect to a Pólya-Gamma (PG) distribution (Polson et al., 2013) as $\omega_i \sim \mathrm{PG}(1,0)$, and hence we have an augmented likelihood as

$$
P(y_i,\omega_i\,|\,\boldsymbol{x}_i,\boldsymbol{\beta}) \propto e^{(y-\frac{1}{2})\boldsymbol{x}_i'\boldsymbol{\beta}-\frac{1}{2}\omega_i(\boldsymbol{x}_i'\boldsymbol{\beta})^2}.
$$

### B.1. Gibbs Sampling via Data Augmentation

Denoting $\mathbf{X} = (\boldsymbol{x}_1, \ldots, \boldsymbol{x}_N)'$, $\boldsymbol{y} = (y_1, \ldots, y_N)'$, $\mathbf{A} = \mathrm{diag}(\alpha_0, \ldots, \alpha_V)'$, and $\mathbf{\Omega} = \mathrm{diag}(\omega_1, \ldots, \omega_N)$, we have

$$(\omega_i \,|\, -) \sim \mathrm{PG}(1, \boldsymbol{x}_i'\boldsymbol{\beta}), \;\; (\boldsymbol{\beta} \,|\, -) \sim \mathcal{N}(\boldsymbol{\mu}, \mathbf{\Sigma}),$$

where $\mathbf{\Sigma} = (\mathbf{A} + \mathbf{X}'\mathbf{\Omega}\mathbf{X})^{-1}$ and $\boldsymbol{\mu} = \mathbf{\Sigma}\mathbf{X}'(\boldsymbol{y} - 1/2)$. To sample from the Pólya-Gamma distribution, a random sample from which can be generated as a weighted sum of an infinite number of $iid$ gamma random variables, we follow Zhou (2016) to truncate the infinite sum to the summation of $M$ gamma random variables, where the parameters of the $M$th gamma random variable are adjusted to match the mean and variance of the finite sum with those of the infinite sum. We set $M = 5$ in this paper.

### B.2. Mean-Field Variational Inference with Diagonal Covariance Matrix

We choose a fully factorized $Q$ distribution as

$$Q = \left[\prod\nolimits_i q(\omega_i)\right]\left[\prod\nolimits_v q(\beta_v)\right].$$

To exploit conjugacy, defining

$$q(\omega_i) = \mathrm{PG}\left(1, \lambda_i\right), \quad q(\beta_v) = \mathcal{N}(\mu_v, \sigma_v^2),$$

we have closed-form coordinate ascent variational inference update equations as

$$\lambda_i = \sqrt{\mathbb{E}[(\boldsymbol{x}_i'\boldsymbol{\beta})^2]} = \sqrt{\boldsymbol{x}_i'\mathbb{E}[\boldsymbol{\beta}\boldsymbol{\beta}']\boldsymbol{x}_i},$$
$$\sigma_v^2 = \left(\mathbb{E}[\alpha_v] + \sum\nolimits_i \mathbb{E}[\omega_i]x_{iv}^2\right)^{-1}$$
$$\mu_v = \sigma_v^2 \sum\nolimits_i x_{iv}\left\{y_i - 1/2 - \mathbb{E}[\omega_i]\sum\nolimits_{\tilde{v}\neq v} x_{i\tilde{v}}\mathbb{E}[\beta_{\tilde{v}}]\right\},$$

where the expectations with respect to the $q$ distributions can be expressed as $\mathbb{E}[\boldsymbol{\beta}\boldsymbol{\beta}'] = \boldsymbol{\mu}\boldsymbol{\mu}' + \mathrm{diag}(\sigma_0^2, \ldots, \sigma_V^2)$ and $\mathbb{E}[\omega_i] = \tanh(\lambda_i/2)/(2\lambda_i)$.

### B.3. Mean-Field Variational Inference with Full Covariance Matrix

We choose a fully factorized $Q$ distribution as

$$Q = \left[\prod\nolimits_i q(\omega_i)\right] q(\boldsymbol{\beta}).$$

To exploit conjugacy, defining

$$q(\omega_i) = \mathrm{PG}\left(1, \lambda_i\right), \quad q(\boldsymbol{\beta}) = \mathcal{N}(\boldsymbol{\mu}, \mathbf{\Sigma}),$$

we have closed-form coordinate ascent variational inference update equations as

$$\lambda_i = \sqrt{\mathbb{E}[(\boldsymbol{x}_i'\boldsymbol{\beta})^2]} = \sqrt{\boldsymbol{x}_i'\mathbb{E}[\boldsymbol{\beta}\boldsymbol{\beta}']\boldsymbol{x}_i},$$
$$\mathbf{\Sigma} = (\mathbb{E}[\mathbf{A}] + \mathbf{X}'\mathbb{E}[\mathbf{\Omega}]\mathbf{X})^{-1}, \; \boldsymbol{\mu} = \mathbf{\Sigma}\mathbf{X}'(\boldsymbol{y} - 1/2),$$

where the expectations with respect to the $q$ distributions can be expressed as $\mathbb{E}[\boldsymbol{\beta}\boldsymbol{\beta}'] = \boldsymbol{\mu}\boldsymbol{\mu}' + \mathbf{\Sigma}$ and $\mathbb{E}[\omega_i] = \tanh(\lambda_i/2)/(2\lambda_i)$. Note the update equations shown above are identical to those shown in Jaakkola & Jordan (2000).

### B.4. SIVI Configuration

For inputs in Algorithm 1, we choose a multi-layer perceptron with layer size $[100, 200, 100]$ as $T_{\boldsymbol{\phi}}$ for $\boldsymbol{\psi} = T_{\boldsymbol{\phi}}(\boldsymbol{\epsilon})$, $\boldsymbol{\epsilon}$ as 50 dimensional isotropic Gaussian random variable and $K = 100$, $J = 50$. For the explicit layer, we choose an MVN as $q_{\boldsymbol{\xi}}(\boldsymbol{z} \,|\, \boldsymbol{\psi}) = \mathcal{N}(\boldsymbol{z}; \boldsymbol{\psi}, \boldsymbol{\xi})$. In this setting, $\boldsymbol{\psi}$ is the mean variable mixed with implicit distribution $q_{\boldsymbol{\phi}}(\boldsymbol{\psi})$ while $\boldsymbol{\xi}$ is the covariance matrix which can be either diagonal or full. In the experiments, we update the neural network parameter $\boldsymbol{\phi}$ by the Adam optimizer, with learning rate 0.01. We update $\boldsymbol{\xi}$ by gradient ascent, with step size $\eta_t = 0.001 * 0.9^{\text{iteration}/100}$. The implicit layer parameter $\boldsymbol{\phi}$ and explicit layer parameter $\boldsymbol{\xi}$ are updated iteratively.

## C. Experimental Settings and Results for SIVAE

We implement SIVI with $M = 3$ stochastic hidden layers, with the dimensions of hidden layers $[\boldsymbol{\ell}_1, \boldsymbol{\ell}_2, \boldsymbol{\ell}_3]$ as $[150, 150, 150]$ and with the dimensions of injected noises $[\boldsymbol{\epsilon}_1, \boldsymbol{\epsilon}_2, \boldsymbol{\epsilon}_3]$ as $[150, 100, 50]$. Between two adjacent stochastic layers there is a fully connected deterministic layer with size 500 and *ReLU* activation function. We choose binary pepper and salt noise (Im et al., 2017) for $q_t(\boldsymbol{\epsilon})$. The model is trained for 2000 epochs with mini-batch size 200 and step-size $0.001 * 0.75^{\text{epoch}/100}$. $K_t$ is gradually increased from 1 to 100 during the first 1500 epochs. The explicit and implicit layers are trained iteratively. Warm-up is used during the first 300 epochs as suggested by Sønderby et al. (2016) to gradually impose the prior regularization term $\mathrm{KL}(q_{\boldsymbol{\phi}}(\boldsymbol{z} \,|\, \boldsymbol{x})||p(\boldsymbol{z}))$. The model is trained end-to-end using the Adam optimizer. After training process, as in Rezende et al. (2014) and Burda et al. (2015), we compute the marginal likelihood for test set by importance sampling with $S = 2000$:

$$\log p(\boldsymbol{x}) \approx \log \frac{1}{S}\sum_{s=1}^{S} \frac{p(\boldsymbol{x} \,|\, \boldsymbol{z}_s)p(\boldsymbol{z}_s)}{\hat{h}(\boldsymbol{z}_s \,|\, \boldsymbol{x})}, \;\; \boldsymbol{z}_s \sim h(\boldsymbol{z}_s|\boldsymbol{x}),$$

where

$$\hat{h}(\boldsymbol{z}_s|\boldsymbol{x}) = \frac{1}{M}\sum_{k=1}^{M} q(\boldsymbol{z}_s \,|\, \boldsymbol{\psi}^{(k)}), \;\; \boldsymbol{\psi}^{(k)} \overset{iid}{\sim} q_{\boldsymbol{\phi}}(\boldsymbol{\psi}|\boldsymbol{x})$$

is used to estimate $h(\boldsymbol{z}_s \,|\, \boldsymbol{x})$; we set $M = 100$. The performance of SIVI and a comparison to reported results with other methods are provided in Table 2.

*Table 2.* Comparison of the negative log evidence between various algorithms.

| **Methods** | $-\log p(\boldsymbol{x})$ |
|---|---|
| *Results below form Burda et al. (2015)* | |
| VAE + IWAE | $= 86.76$ |
| IWAE + IWAE | $= 84.78$ |
| *Results below form Salimans et al. (2015)* | |
| DLGM + HVI (1 leapfrog step) | $= 88.08$ |
| DLGM + HVI (4 leapfrog step) | $= 86.40$ |
| DLGM + HVI (8 leapfrog steps) | $= 85.51$ |
| *Results below form Rezende & Mohamed (2015)* | |
| DLGM+NICE (Dinh et al., 2014) (k = 80) | $\leq 87.2$ |
| DLGM+NF (k = 40) | $\leq 85.7$ |
| DLGM+NF (k = 80) | $\leq 85.1$ |
| *Results below form Gregor et al. (2015)* | |
| DLGM | $\approx 86.60$ |
| NADE | $= 88.33$ |
| DBM 2hl | $\approx 84.62$ |
| DBN 2hl | $\approx 84.55$ |
| EoNADE-5 2hl (128 orderings) | $= 84.68$ |
| DARN 1hl | $\approx 84.13$ |
| *Results below form Maaløe et al. (2016)* | |
| Auxiliary VAE (L=1, IW=1) | $\leq 84.59$ |
| *Results below form Mescheder et al. (2017)* | |
| VAE + IAF (Kingma et al., 2016) | $\approx 84.9 \pm 0.3$ |
| Auxiliary VAE (Maaløe et al., 2016) | $\approx 83.8 \pm 0.3$ |
| AVB + AC | $\approx 83.7 \pm 0.3$ |
| SIVI (3 stochastic layers) | $= 84.07$ |
| SIVI (3 stochastic layers)+ IW($\tilde{K} = 10$) | $= 83.25$ |

## D. Additional Figures

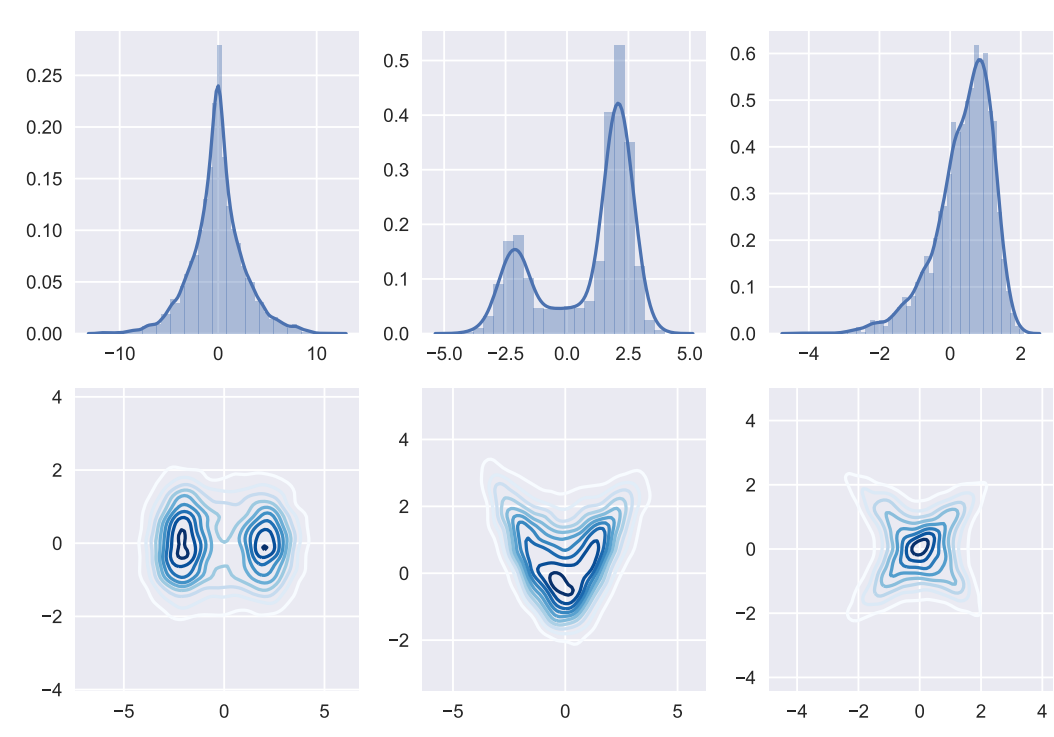

*Figure 8.* Visualization of the MLP based implicit distributions $\boldsymbol{\psi} \sim q(\boldsymbol{\psi})$, which are mixed with isotropic Gaussian (or Log-Normal) distributions to approximate the target distributions shown in Figure 1.

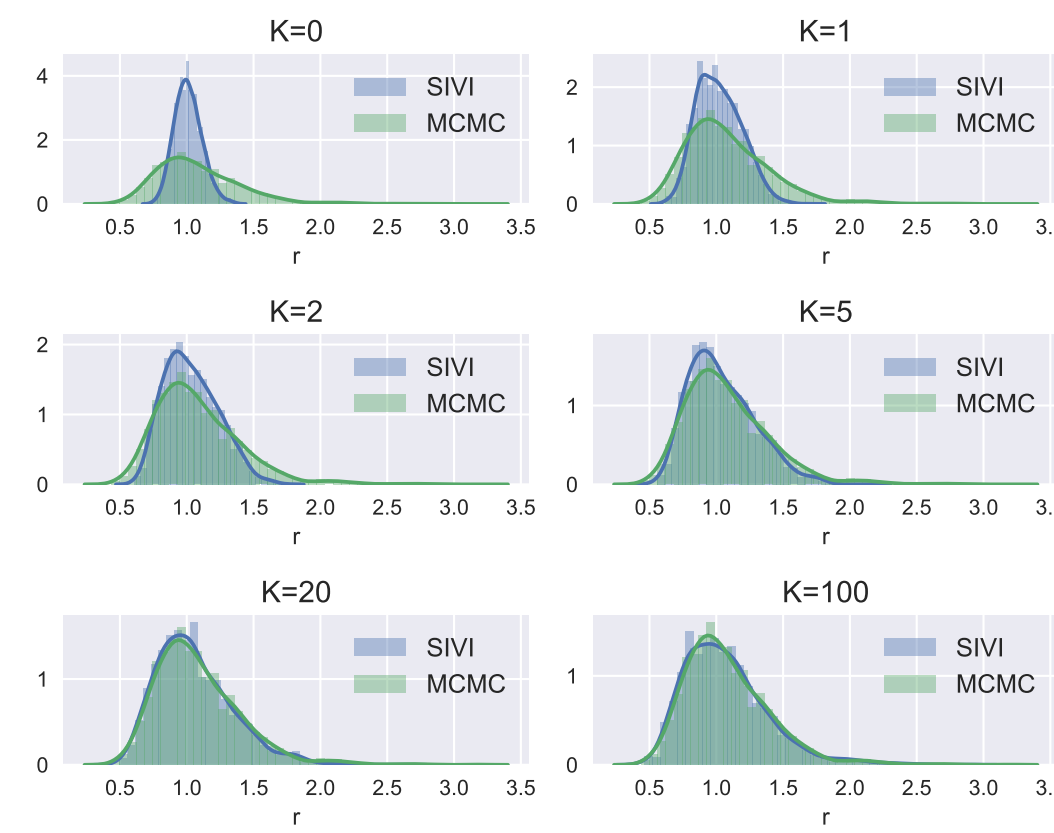

*Figure 9.* The marginal posterior distribution of the negative binomial dispersion parameter $r$ inferred by SIVI becomes more accurate as $K$ increases

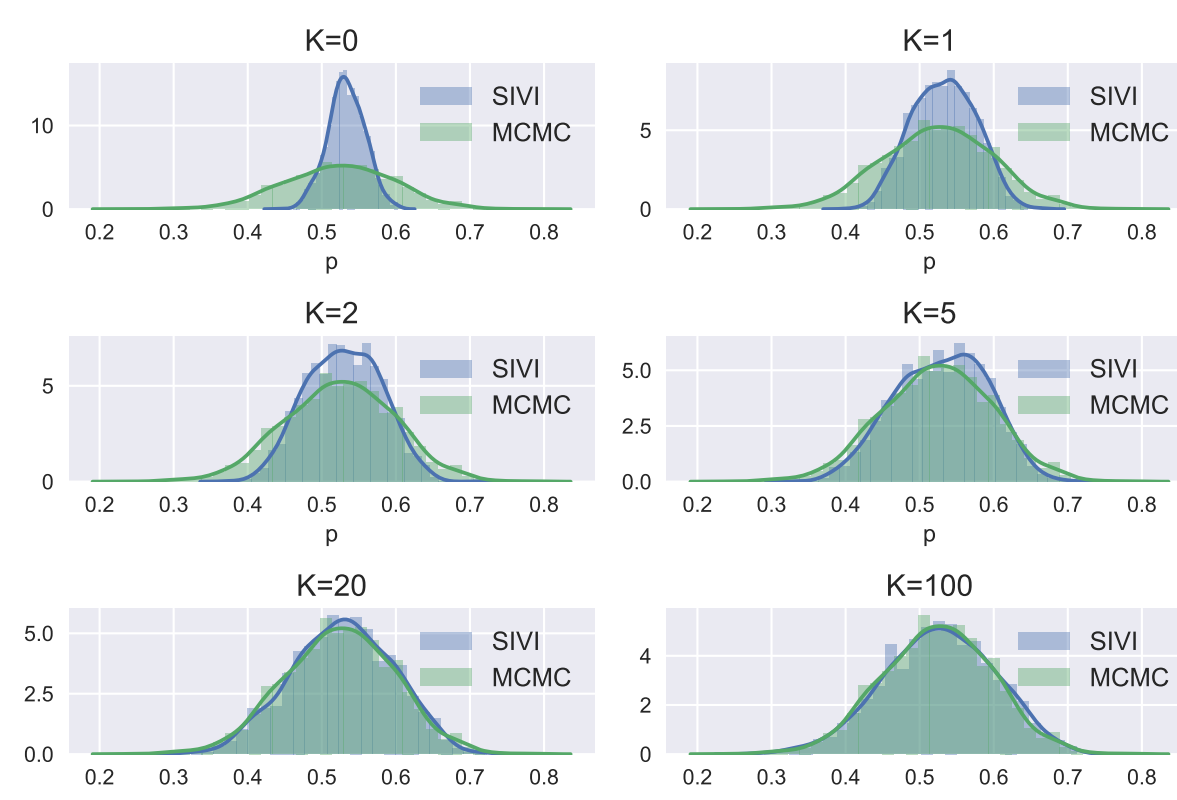

*Figure 10.* The marginal posterior distribution of the negative binomial probability parameter $p$ inferred by SIVI becomes more accurate as $K$ increases.

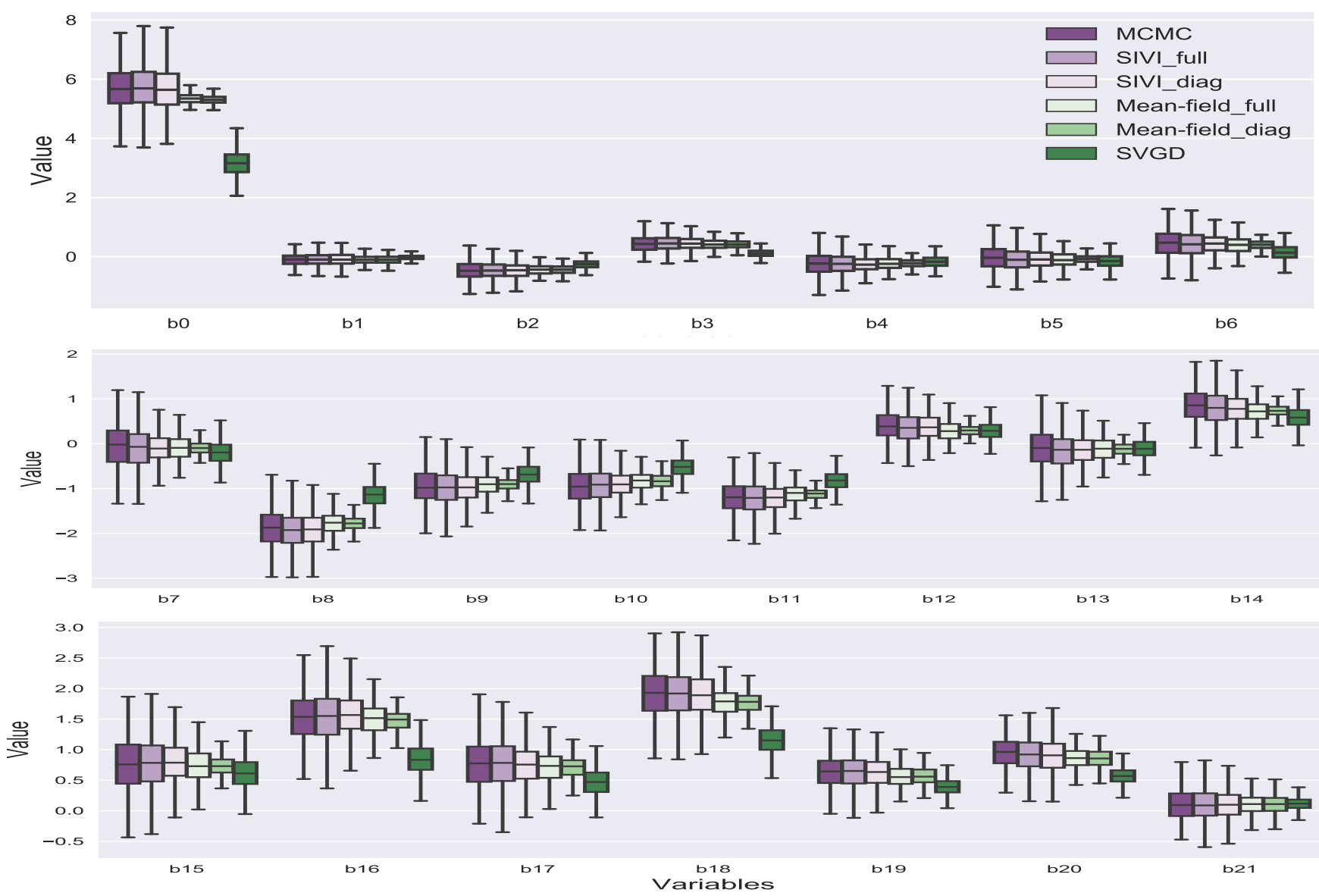

*Figure 11.* Comparison of all marginal posteriors of $\beta_v$ inferred by various methods for Bayesian logistic regression on *waveform*.

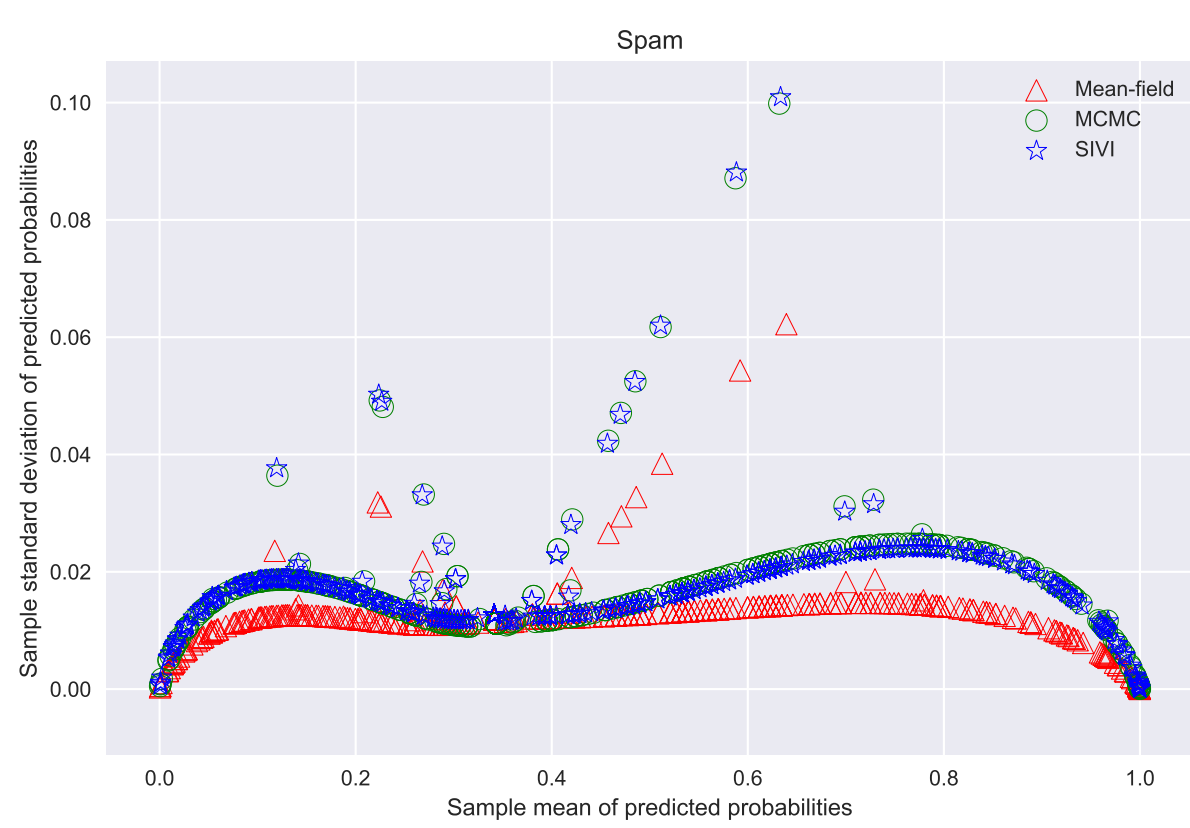

*Figure 12.* Sample means and standard deviations of predictive probabilities for dataset *spam*.

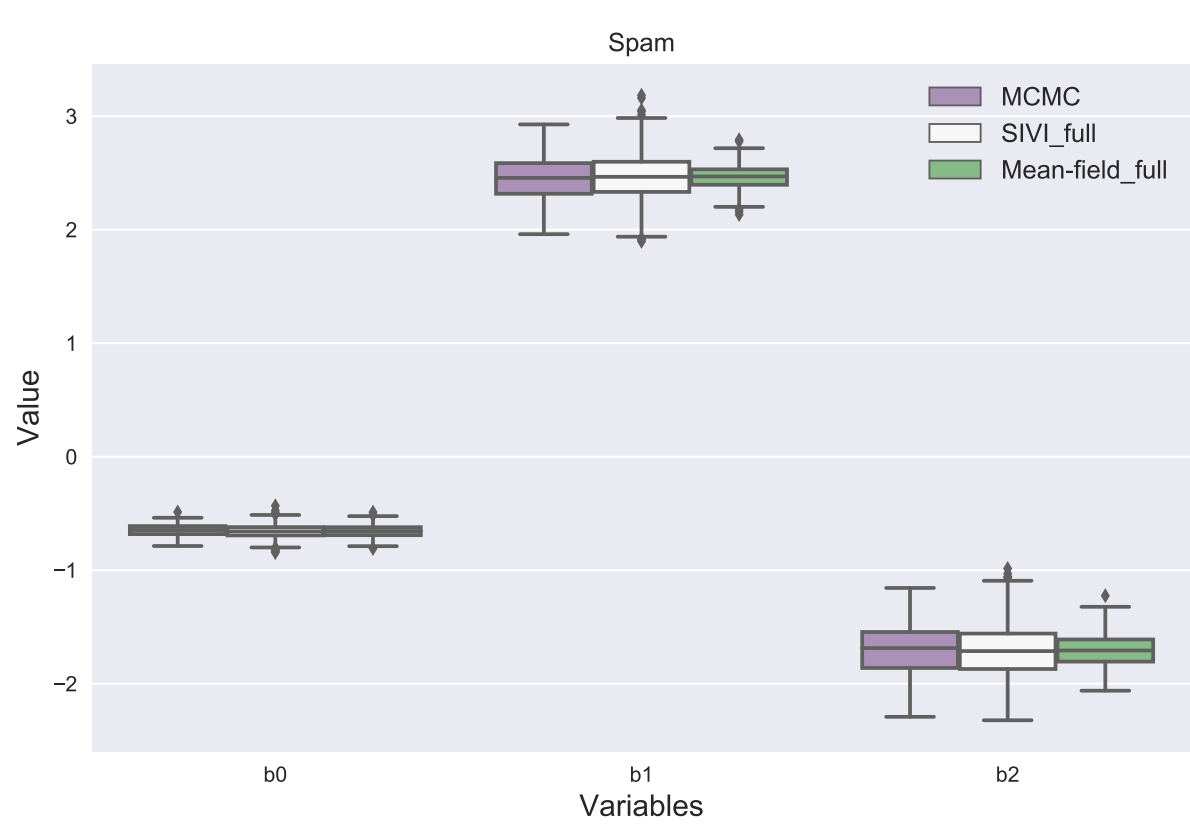

*Figure 13.* Boxplot of marginal posteriors inferred by MCMC, SIVI, and MFVI for dataset *spam*.

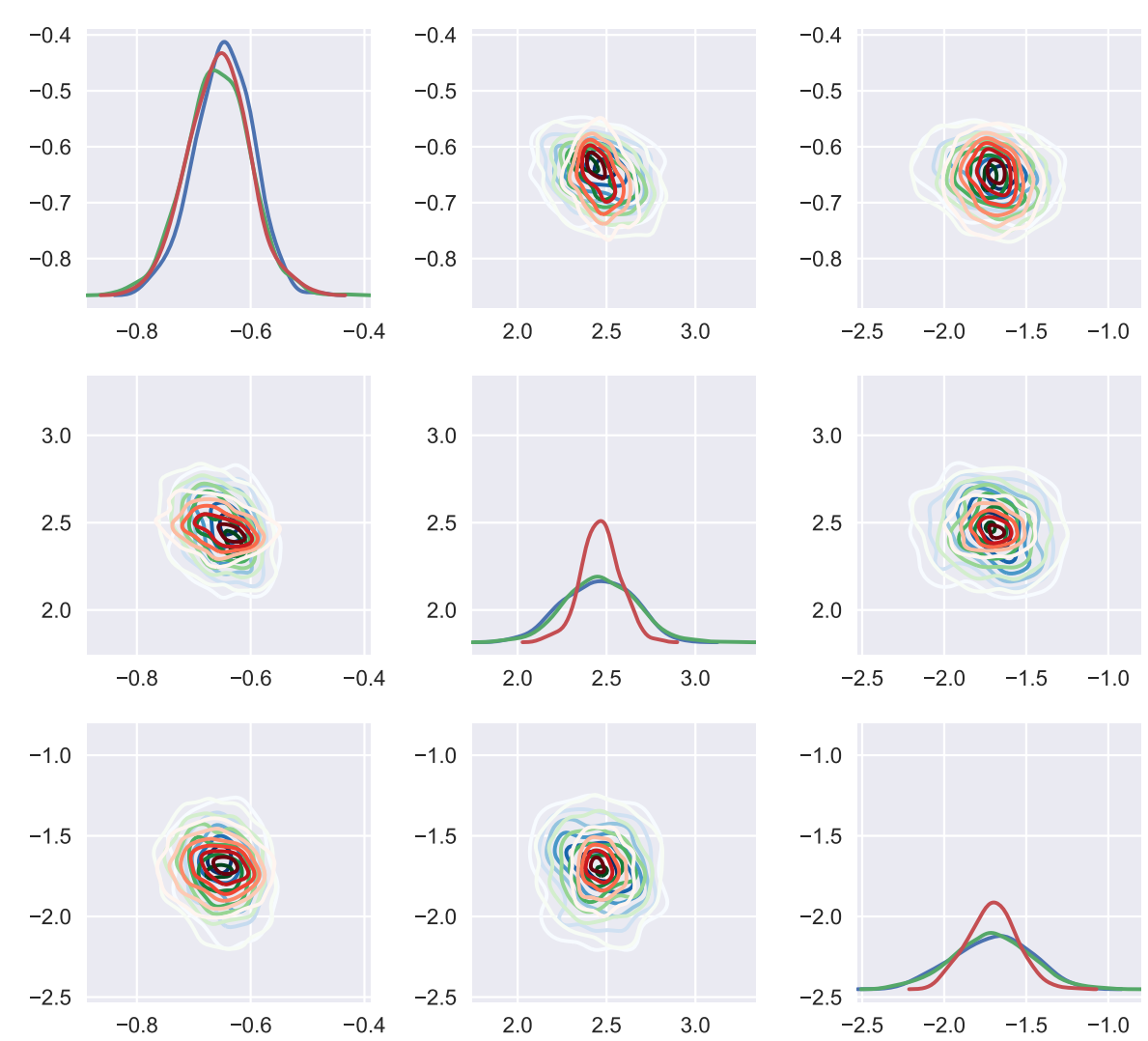

*Figure 14.* Univariate marginal and pairwise joint posteriors for dataset *spam*. Blue, green, and red are for MCMC, SIVI with a full covariance matrix, and MFVI with a full covariance matrix, respectively.

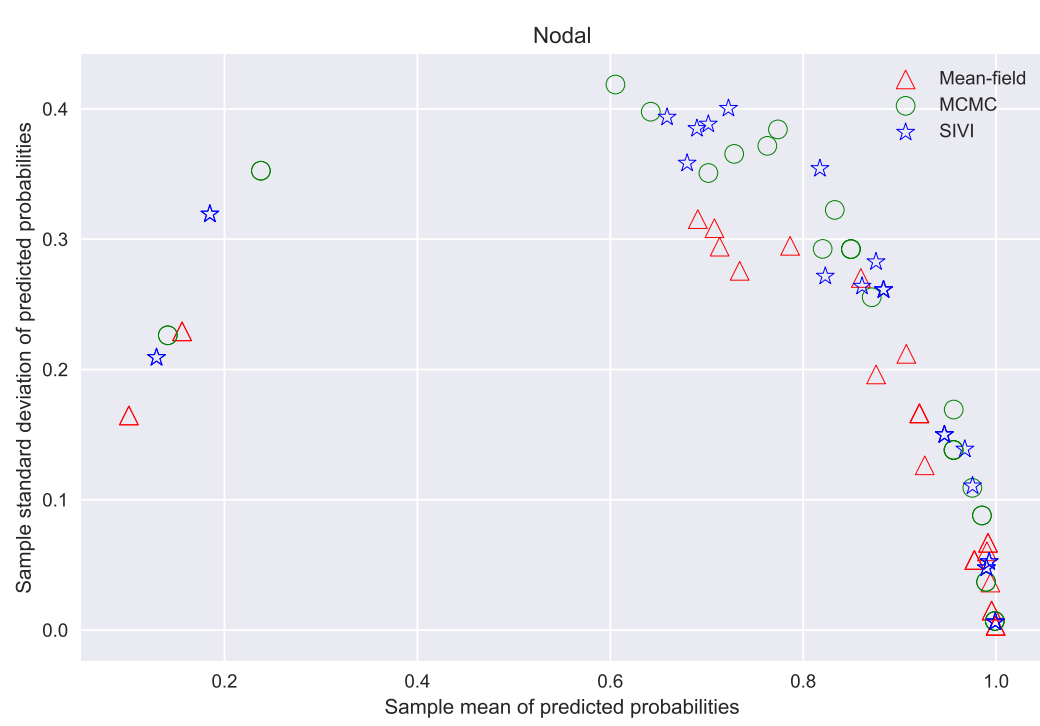

*Figure 15.* Sample means and standard deviations of predictive probabilities for dataset *nodal*.

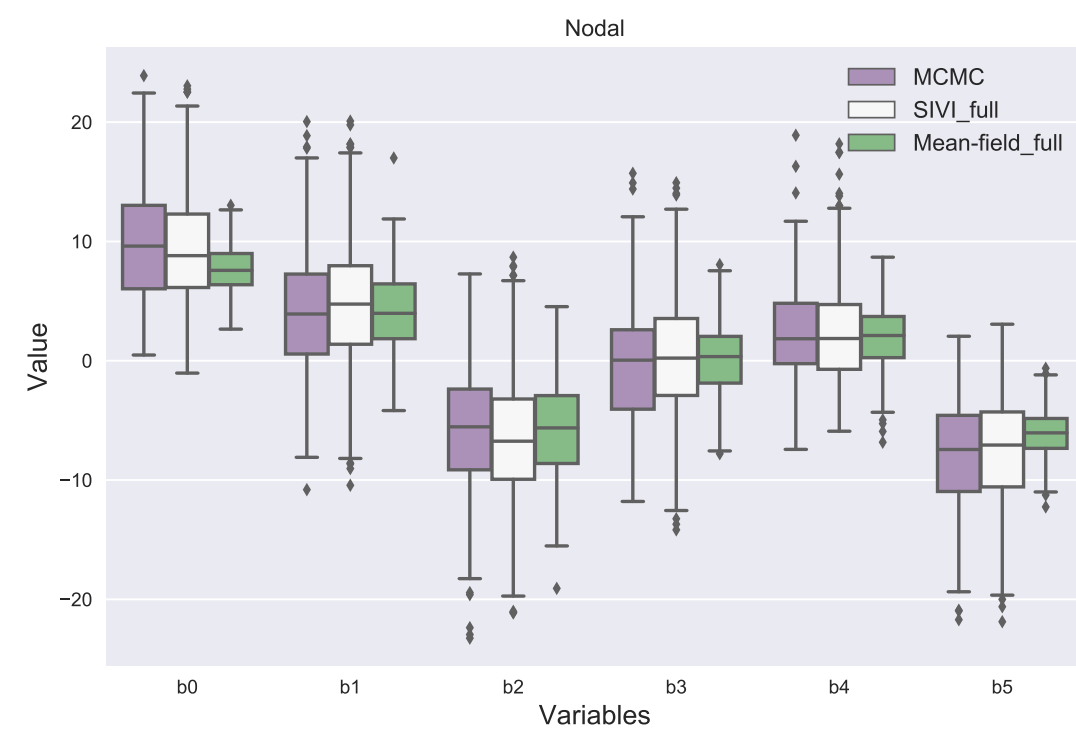

*Figure 16.* Boxplot of marginal posteriors inferred by MCMC, SIVI, and MFVI for dataset *nodal*.

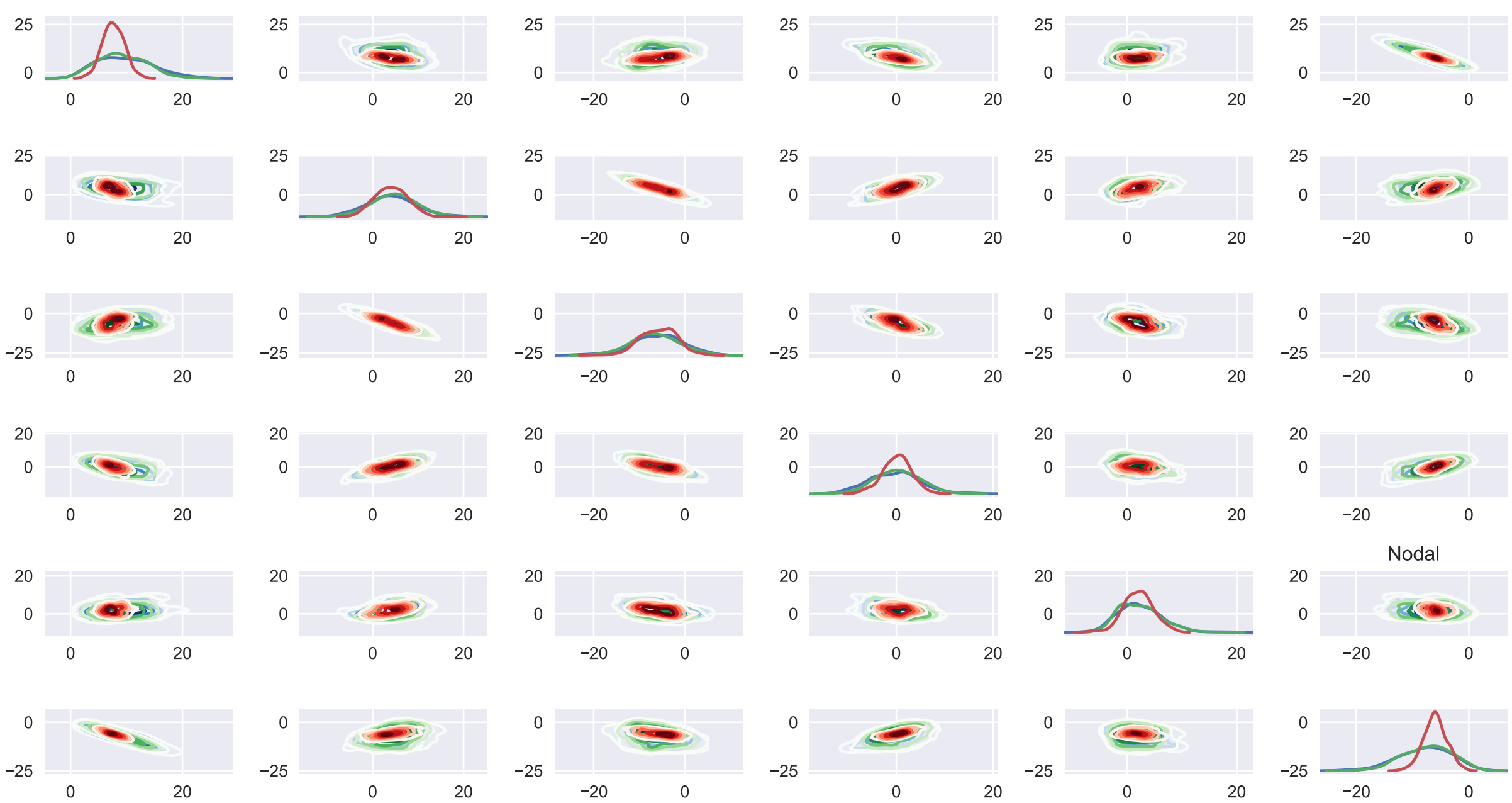

*Figure 17.* Univariate marginal and pairwise joint posteriors for dataset *nodal*. Blue, green, and red are for MCMC, SIVI with a full covariance matrix, and MFVI with a full covariance matrix, respectively.